\documentclass{article}
\usepackage[preprint]{neurips_2019}

\usepackage[utf8]{inputenc} %
\usepackage[T1]{fontenc}    %
\usepackage{hyperref}       %
\usepackage{url}            %
\usepackage{booktabs}       %
\usepackage{amsfonts}       %
\usepackage{nicefrac}       %
\usepackage{microtype}      %
\usepackage{times}          %
\usepackage{natbib}
\usepackage{amsmath,amssymb}
\usepackage[makeroom]{cancel}
\usepackage{enumitem}
\usepackage{xargs}  
\usepackage{soul}
\usepackage{lipsum} 
\usepackage{xcolor}
\usepackage[colorinlistoftodos,textsize=tiny]{todonotes}
\usepackage{chngcntr}
\usepackage{subcaption}
\usepackage{wrapfig}
\usepackage{xfrac}
\usepackage{multirow}
\usepackage{defs}

\makeatletter
\def\thanks#1{\protected@xdef\@thanks{\@thanks
        \protect\footnotetext{#1}}}
\makeatother

\setlist[itemize]{noitemsep,topsep=-6pt}
\setlist[enumerate]{noitemsep,topsep=-6pt}

\title{Neural Multisensory Scene Inference}
\author{
Jae Hyun Lim$^{\dagger\mathsection 123}$, Pedro O. Pinheiro$^{1}$, 
\textbf{Negar Rostamzadeh}$^1$,\\ \textbf{Christopher Pal}$^{1234}$\textbf{,} \textbf{Sungjin Ahn}$^{\ddagger\mathsection 5}$
\\
$^1$Element AI, $^2$Mila, $^3$Universit\'e de Montr\'eal, $^4$Polytechnique Montr\'eal, $^5$Rutgers University
\thanks{$^\dagger$Work done during the internship of JHL at Element AI.
$^\ddagger$Part of the work had done while SA was at Element AI.
$^\mathsection$Correspondence to \href{mailto:jae.hyun.lim@umontreal.ca}{jae.hyun.lim@umontreal.ca} and \href{mailto:sungjin.ahn@cs.rutgers.edu}{sungjin.ahn@cs.rutgers.edu}
}
}

\begin{document}
\maketitle
\begin{abstract}
For embodied agents to infer representations of the underlying 3D physical world they inhabit, they should efficiently combine multisensory cues from numerous trials, e.g., by looking at and touching objects.~Despite its importance, multisensory 3D scene representation learning has received less attention compared to the unimodal setting.~In this paper, we propose the \emph{Generative Multisensory Network} (GMN) for learning latent representations of 3D scenes which are partially observable through multiple sensory modalities.~We also introduce a novel method, called the \emph{Amortized Product-of-Experts}, to improve the computational efficiency and the robustness to unseen combinations of modalities at test time.~Experimental results demonstrate that the proposed model can efficiently infer robust modality-invariant 3D-scene representations from arbitrary combinations of modalities and perform accurate cross-modal generation.~To perform this exploration, we also develop the Multisensory Embodied 3D-Scene Environment (MESE).
\end{abstract}

\section{Introduction}
Learning a world model and its representation is an effective way of solving many challenging problems in machine learning and robotics, \emph{e.g.}, via model-based reinforcement learning~\citep{silver2016mastering}.
One characteristic aspect in learning the physical world is that it is inherently multifaceted and that we can perceive its complete characteristics only through our multisensory modalities.~Thus, incorporating different physical aspects of the world via different modalities should help build a richer model and representation. One approach to learn such multisensory representations is to learn a \emph{modality-invariant} representation as an abstract concept representation of the world. This is an idea well supported in both psychology and neuroscience.~According to the \emph{grounded cognition} perspective \citep{barsalou2008grounded}, such abstract concepts like objects and events can only be obtained through perceptual signals.~For example, what represents a cup in our brain is its visual appearance, the sound it could make, the tactile sensation, etc.~In neurosciences, the existence of \emph{concept cells}~\citep{quiroga2012concept} that responds only to a specific concept regardless of the modality sourcing the concept (\emph{e.g.}, by showing a picture of Jennifer Aniston or listening her name) can be considered as a biological evidence of the \emph{metamodal} brain perspective~\citep{pascual2001metamodal, yildirim2014perception} and the modality-invariant representation.

An unanswered question from the computational perspective (our particular interest in this paper) is how to learn such modality-invariant representation of the complex physical world (\emph{e.g.,} 3D scenes placed with objects).
~We argue that it is a particularly challenging problem because the following requirements need to be satisfied for the learned world model.
First, the learned representation should reflect the 3D nature of the world.~Although there have been some efforts in learning multimodal representations (see Section~\ref{sc:related_works}), those works do not consider this fundamental 3D aspect of the physical world.~Second, the learned representation should also be able to model the intrinsic stochasticity of the world.~Third, for the learned representation to generalize, be robust, and to be practical in many applications, the representation should be able to be inferred from experiences of any partial combinations of modalities. It should also facilitate the generative modelling of other arbitrary combinations of modalities~\citep{yildirim2014perception}, supporting the metamodal brain hypothesis -- for which human evidence can be found from the phantom limb phenomenon~\citep{ramachandran1998perception}.~Fourth, even if it is evidenced that there exists metamodal representation, there still exist modality-dependent brain regions, revealing the modal-to-metamodal hierarchical structure~\citep{rohe2016distinct}. A learning model can also benefit from such hierarchical representation as shown by~\cite{HsuG18}.~Lastly, the learning should be computationally efficient and scalable, \emph{e.g.}, with respect to the number of possible modalities.

Motivated by the above desiderata, we propose the Generative Multisensory Network (GMN) for neural multisensory scene inference and rendering.~In GMN, from an arbitrary set of source modalities we infer a 3D representation of a scene that can be queried for generation via an arbitrary target modality set, a property we call \textit{generalized cross-modal generation}. To this end, we formalize the problem as a probabilistic latent variable model based on the Generative Query Network \citep{Eslami1204/science18} framework and introduce the Amortized Product-of-Experts (APoE). The prior and the posterior approximation using APoE makes the model trainable only with a small combinations of modalities, instead of the entire combination set.~The APoE also resolves the inherent space complexity problem of the traditional Product-of-Experts model and also improves computation efficiency. As a result, the APoE allows the model to learn from a large number of modalities without tight coupling among the modalities, a desired property in many applications such as Cloud Robotics \citep{saha2018comprehensive} and Federated Learning \citep{fedlearn16}.~In addition, with the APoE the modal-to-metamodal hierarchical structure is easily obtained. In  experiments, we show the above properties of the proposed model on 3D scenes with blocks of various shapes and colors along with a human-like hand.

The contributions of the paper are as follows:
(i) We introduce a formalization of modality-invariant multisensory 3D representation learning using a generative query network model and propose the Generative Multisensory Network (GMN)\footnote{Code is available at: \url{https://github.com/lim0606/pytorch-generative-multisensory-network}}.
(ii) We introduce the Amortized Product-of-Experts network that allows for generalized cross-modal generation while resolving the problems in the GQN and traditional Product-of-Experts.
(iii) Our model is the first to extend multisensory representation learning to 3D scene understanding with human-like sensory modalities (such as haptic information) and cross-modal generation. 
(iv) We also develop the Multisensory Embodied 3D-Scene Environment (MESE) used to develop and test the model.

\section{Neural Multisensory Scene Inference}

\subsection{Problem Description}
\label{sec:problem-description}
\begin{figure}[!tb]
\centering
\includegraphics[width=12cm]{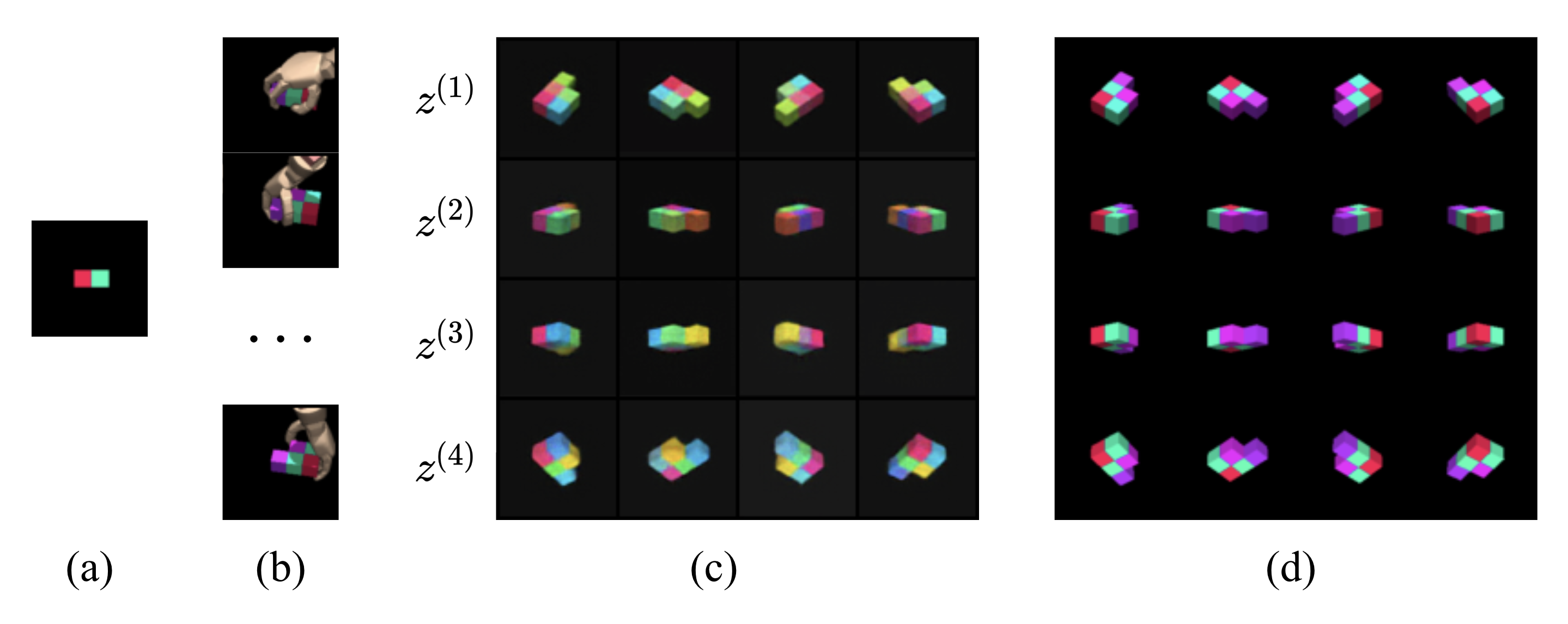}
\vspace{-0.2cm}
\caption{
Cross-modal inference using scene representation.
\small
(a) A single image context.
(b) Haptic contexts. (c) Generated images for some viewpoints (image queries) in the scene, given the contexts. (d) Ground truth images for the same queries.
Conditioning on an image context and multiple haptic contexts, modality-agnostic latent scene representation, $z$, is inferred.
Given sampled $z$s, images are generated using various queries; in (c), each row corresponds to the same latent sample.
Note that the shapes of predicted objects are consistent given different samples $z^{(i)}$, while color pattern of the object changes except the parts seen by the image context (a).
}
\label{fig:result-cross-modal-generation}
\vspace{-0.5cm}
\end{figure}

Our goal is to understand 3D scenes by learning a \emph{metamodal} representation of the scene through the interaction of multiple sensory modalities such as vision, haptics, and auditory inputs.~In particular, motivated by human multisensory processing~\citep{deneve2004bayesian,shams2008benefits,murray2011neural}, we consider a setting where the model infers a scene from experiences of a set of modalities and then to generate another set of modalities given a query for the generation.~For example, we can experience a 3D scene where a cup is on a table only by touching or grabbing it from some hand poses and ask if we can visually imagine the appearance of the cup from an arbitrary query viewpoint (see Fig. \ref{fig:result-cross-modal-generation}).~We begin this section with a formal definition of this problem.

A multisensory scene, simply a scene, $S$ consists of context $C$ and observation $O$. Given the set of all available modalities $\cM$, the context and observation in a scene are obtained through the context modalities $\cM_c(S) \subset \cM$ and the observation modalities $\cM_o(S) \subset \cM$, respectively. In the following, we omit the scene index $S$ when the meaning is clear without it. Note that $\cM_c$ and $\cM_o$ are arbitrary subsets of $\cM$ including the cases $\cM_o \cap \cM_c = \emptyset$, $\cM_o = \cM_c$, and $\cM_o \cup \cM_c \subsetneq \cM$. We also use $\cM_S$ to denote all modalities available in a scene, $\cM_o(S)\cup \cM_c(S)$. %

The context and observation consist of sets of experience trials represented as \emph{query($\bv$)-sense($\bx$)} pairs, \emph{i.e.}, $C=\{(\bv_n,\bx_n)\}_{n=1}^{N_c}$ and $O=\{(\bv_n,\bx_n)\}_{n=1}^{N_o}$.
For convenience, we denote the set of queries and senses in observation by $V$ and $X$, respectively, \textit{i.e.}, $O=(V,X)$. 
Each query $\bv_n$ and sense $\bx_n$ in a context consists of \textit{modality-wise} queries and senses corresponding to each modality in the context modalities, \emph{i.e.}, $(\bv_n,\bx_n)=\{(\bv_n^m, \bx_n^m)\}_{m\in\cM_c}$ (See Fig. \ref{fig:multisensory-scene}). %
Similarly, the query and the sense in observation $O$ is constrained to have only the observation modalities $\cM_o$. For example, for modality $m=\tt{vision}$, an unimodal query $\bv_n^{\tt{vision}}$ can be the viewpoint and the sense $\bx_n^{\tt{vision}}$ is the observation image obtained from the query viewpoint. Similarly, for $m=\tt{haptics}$, an unimodal query $\bv_n^{\tt{haptics}}$ can be the hand position, and the sense $\bx_n^{\tt{haptics}}$ is the tactile and pressure senses obtained by a grab from the query hand pose. For a scene, we may have $\cM_c = \{\tt{haptics}, \tt{auditory}\}$ and $\cM_o = \{\tt{vision}, \tt{auditory}\}$. For convenience, we also introduce the following notations. We denote the context corresponding only to a particular modality $m$ by $C_m=\{(\bv_n^m,\bx_n^m)\}_{n=1}^{N_c^m}$ such that $N_c = \sum_m N_c^m$ and  $C=\{C_m\}_{m\in\cM_c}$. 
Similarly, $O_m$, $X_m$ and $V_m$ are used to denote modality $m$ part of $O$, $X$, and $V$, respectively. 

Given the above definitions, we formalize the problem as learning a generative model of a scene that can generate senses $X$ corresponding to queries $V$ of a set of modalities, provided a context $C$ from other arbitrary modalities. Given scenes from the scene distribution $(O,C)\sim P(S)$, our training objective is to maximize
$
    \eE_{(O,C)\sim P(S)}[\log P_\ta(X|V,C)]
    \label{eq:obj}
$,
where $\ta$ is the model parameters to be learned.
\subsection{Generative Process} 
We formulate this problem as a probabilistic latent variable model where we introduce the latent metamodal scene representation $\bz$ from a conditional prior $P_\ta(\bz|C)$. The joint distribution of the generative process becomes:
\eq{
    P_\ta(X,\bz|V,C) 
    &= P_\ta(X|V,\bz)P_\ta(\bz|C)\nn\\
    &= \prod_{n=1}^{N_o}P_\ta(\bx_n|\bv_n,\bz)P_\ta(\bz|C) = \prod_{n=1}^{N_o}\prod_{m\in\cM_o}P_{\ta_m}(\bx_n^m|\bv_n^m,\bz)P_\ta(\bz|C).
}
\subsection{Prior for Multisensory Context}
\label{gen-process-prior-for-multisensory-context}
As the prior $P_\ta(\bz|C)$ is conditioned on the context, we need an encoding mechanism of the context to obtain $\bz$.~A simple way to do this is to follow the Generative Query Network (GQN)~\citep{Eslami1204/science18} approach: each context query-sense pair $(\bv_n,\bx_n)$ is encoded to $\br_n=f_{\text{enc}}(\bv_n,\bx_n)$ and summed (or averaged) to obtain permutation-invariant context representation $\br = \sum_{n} \br_n$. A ConvDRAW module \citep{GregorBRDW16/nips} is then used to sample $\bz$ from $\br$.

In the multisensory setting, however, this approach cannot be directly adopted due to a few challenges.
First, unlike GQN the sense and query of each sensor modality has different structure, and thus we cannot have a single and shared context encoder that deals with all the modalities. In our model, we therefore introduce a modality encoder $\br^m = 
\sum_{(\bx,\bv)\in C_m} f_{\text{enc}}^m(\bx,\bv)$ for each $m \in \cM$. 

The second challenge stems from the fact that we want our model capable of generating from any context modality set $\cM_c(S)$ to any observation modality set $\cM_o(S)$ -- a property we call \textit{generalized cross-modal generation} (GCG). However, at test time we do not know which sensory modal combinations will be given as a context and a target to generate. This hence requires collecting a training data that contains all possible combinations of context-observation modalities $\cM^*$. 
This equals the Cartesian product of $\cM$'s powersets, \emph{i.e.}, $\cM^* = Power(\cM)\times Power(\cM)$. This is a very expensive requirement as $|\cM^*|$ increases exponentially with respect to the number of modalities\footnote{The number of modalities or sensory input sources can be very large depending on the application. Even in the case of `human-like' embodied learning, it is not only, vision, haptics, auditory, etc. For example, given a robotic hand, the context input sources can be only a part of the hand, \emph{e.g.}, some parts of some fingers, from which we humans can imagine the senses of other parts.} $|\cM|$.

Although one might consider dropping-out random modalities during training to achieve the generalized cross-modal generation, this still assumes the availability of the full modalities from which to drop off some modalities. Also, it is unrealistic to assume that we always have access to the full modalities; to learn, we humans do not need to touch everything we see. Therefore, it is important to make the model \emph{learnable only with a small subset of all possible modality combinations while still achieving the GCG property}. We call this the \textit{missing-modality} problem.

To this end, we can model the conditional prior as a Product-of-Experts (PoE) network~\citep{Hinton02/neco} with one expert per sensory modality parameterized by $\ta_m$.~That is,
$
P(\bz|C) 
\propto \prod_{m\in\cM_c}P_{\ta_m}(\bz|C_m).
$
While this could achieve our goal at the functional level, it comes at a computational cost of increased space and time complexity w.r.t. the number of modalities. This is particularly problematic when we want to employ diverse sensory modalities (as in, \emph{e.g.}, robotics) or if each expert has to be a powerful (hence expensive both in computation and storage) model like the 3D scene inference task~\citep{Eslami1204/science18}, where it is necessary to use the powerful ConvDraw network to represent the complex 3D scene. 

\subsection{Amortized Product-of-Experts as Metamodal Representation} 
To deal with the limitations of PoE, we introduce the \emph{Amortized Product-of-Experts} (APoE).%
~For each modality $m \in \cM_c$, we first obtain modality-level representation $\br^m$ using the modal-encoder. Note that this modal-encoder is a much lighter module than the full ConvDraw network. Then, each modal-encoding $\br^m$ along with its modality-id $m$ is fed into the expert-amortizer  $P_\psi(\bz|\br^m,m)$ that is shared across all modal experts through shared parameter $\psi$. In our case, this is implemented as a ConvDraw module (see Appendix \ref{sec:appendix-network-architectures} for the implementation details).~We can write the APoE prior as follows:
\eq{
    \label{eq:posterior-APoE}
    P(\bz|C) = \prod_{m\in\cM_c}P_\psi(\bz|\br^m,m)\;.
}
We can extend this further to obtain a hierarchical representation model by treating $\br^m$ as a latent variable:
$$P(\bz,\{\br^m\}|C) \propto \prod_{m\in \cM_c} P_\psi(\bz|\br^m,m)P_{\ta_m}(\br^m|C_m)\;,$$
where $\br^m$ is modality-level representation and $\bz$ is metamodal representation. Although we can train this hierarchical model with reparameterization trick and Monte Carlo sampling, for simplicity in our experiments we use deterministic function for $P_{\ta_m}(\br^m|C_m) = \delta[\br^m = f_{\ta_m}(C_m)]$ where $\delta$ is a dirac delta function. In this hierarchical version, the generative process becomes:
\eq{
    &P(X,\bz, \{\br^m\}|V,C) = P_\ta(X|V,\bz)\prod_{m\in\cM_c} P_\psi(\bz|\br^m,m)P_{\ta_m}(\br^m|C_m)\;.
}
An illustration of the generative process is provided in Fig.\ref{fig:computation-graph-gen} (b), on the Appendix. From the perspective of cognitive psychology, the APoE model can be considered as a computational model of the \emph{metamodal} brain hypothesis~\citep{pascual2001metamodal}, which states the existence of metamodal brain area (the expert-of-experts in our case) which perform a specific function  not specific to input sensory modalities.

\subsection{Inference} 
Since the optimization of the aforementioned objective is intractable, we perform variational inference by maximizing the following evidence lower bound (ELBO), $\cL_S(\ta,\phi)$, with the reparameterization trick~\citep{kingma2013auto,rezende2014stochastic}:
\eq{
    \log P_\ta(X|V,C) \geq \eE_{Q_\phi(\bz | C,O)}\left[\log P_\ta(X|V,\bz) \right] -\KL[Q_\phi(\bz|C,O)|| P_\ta(\bz|C)]\;,
}
where $P_\ta(X|V,\bz)=\pd{n}{N_o}\prod_{m\in\cM_o} P_{\ta_m}(\bx_n^m |\bz,\bv_n^m)$.~This can be considered a cognitively-plausible objective as, according to the ``grounded cognition'' perspective \citep{barsalou2008grounded}, the modality-invariant representation of an abstract concept, $\bz$, can never be fully modality-independent.

\begin{figure}[!tb]
    \centering
    \hspace{-10mm}
    \includegraphics[width=4.3cm]{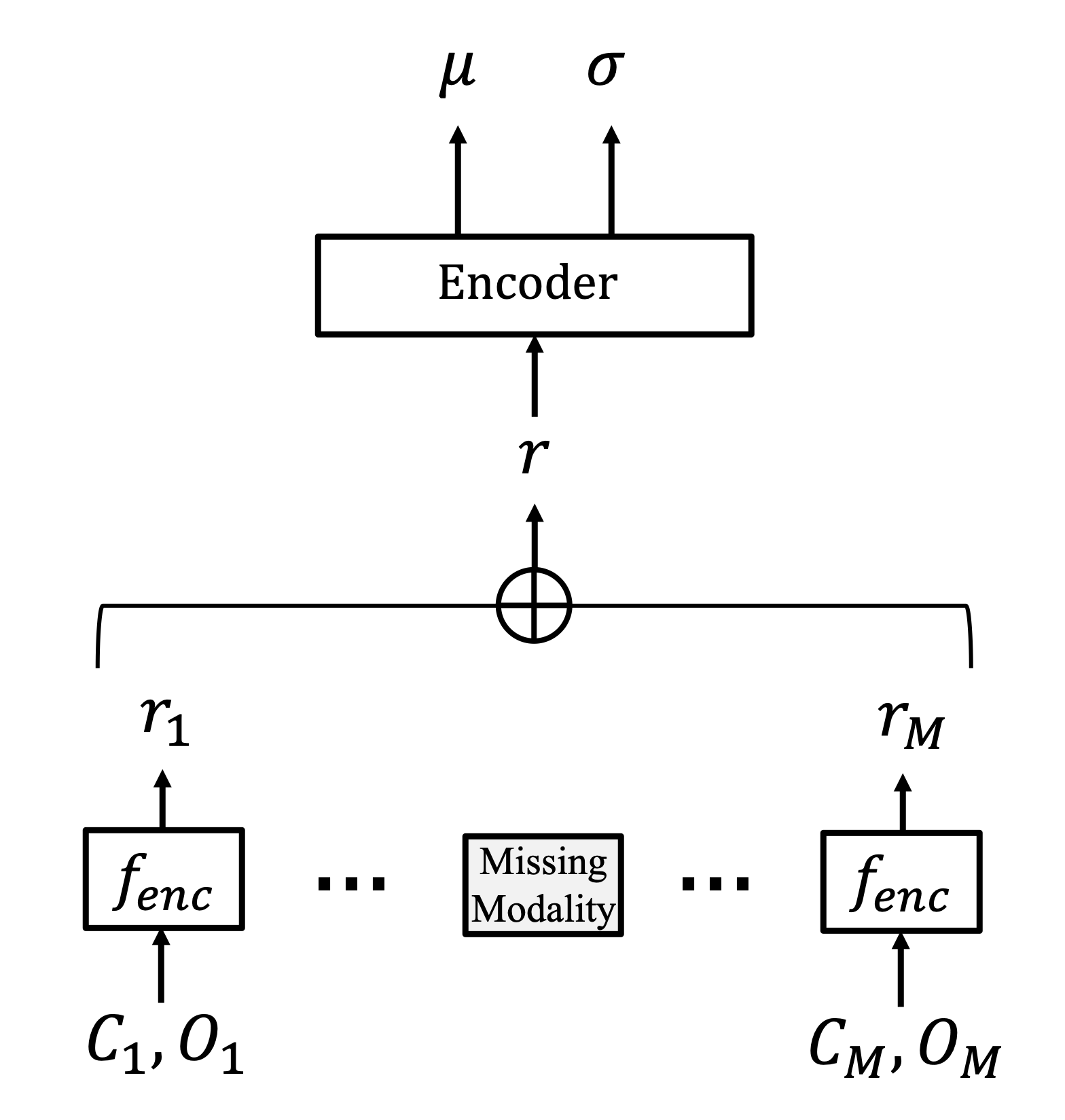}
    \hspace{3mm}
    \includegraphics[width=4.3cm]{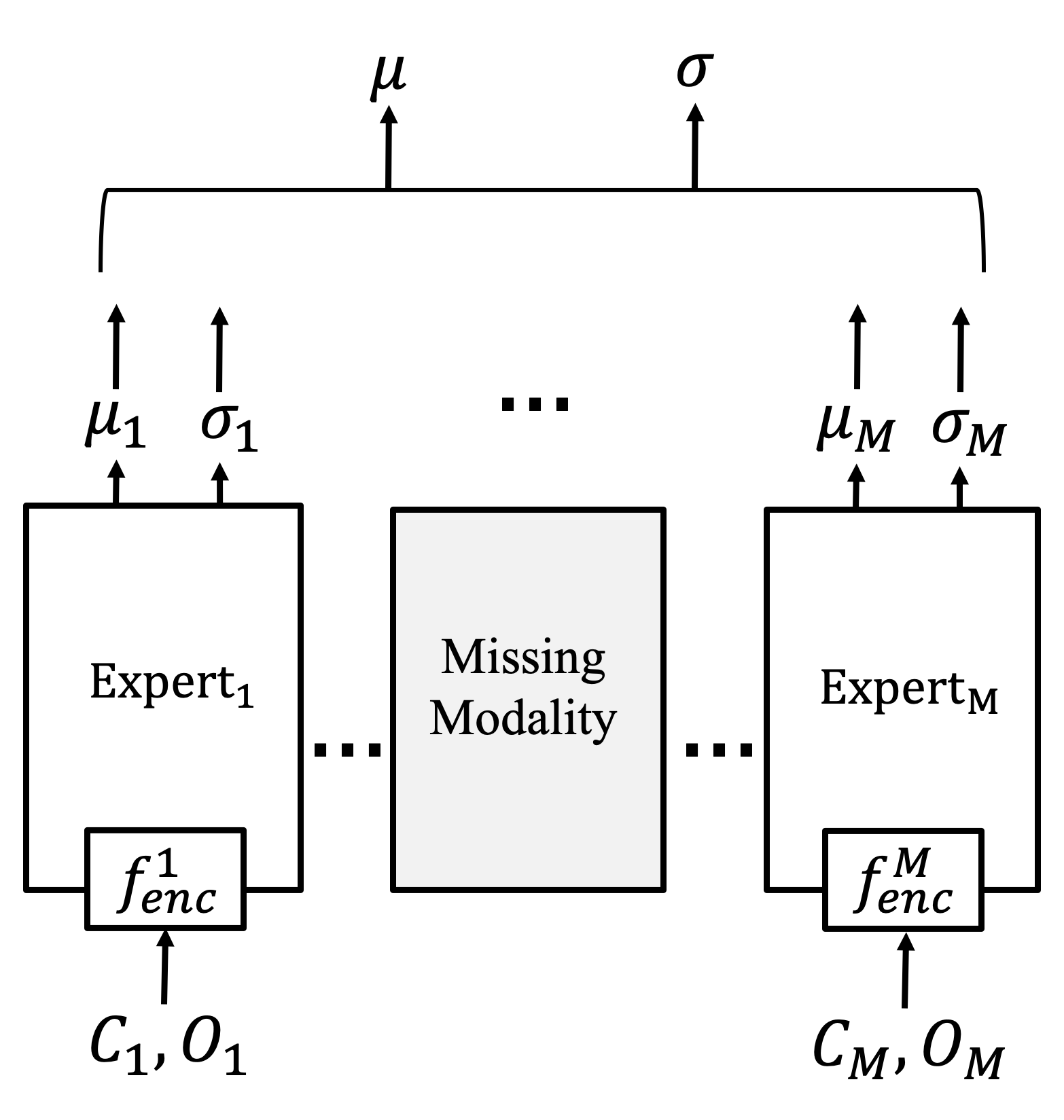}
    \hspace{3mm}
    \includegraphics[width=4.3cm]{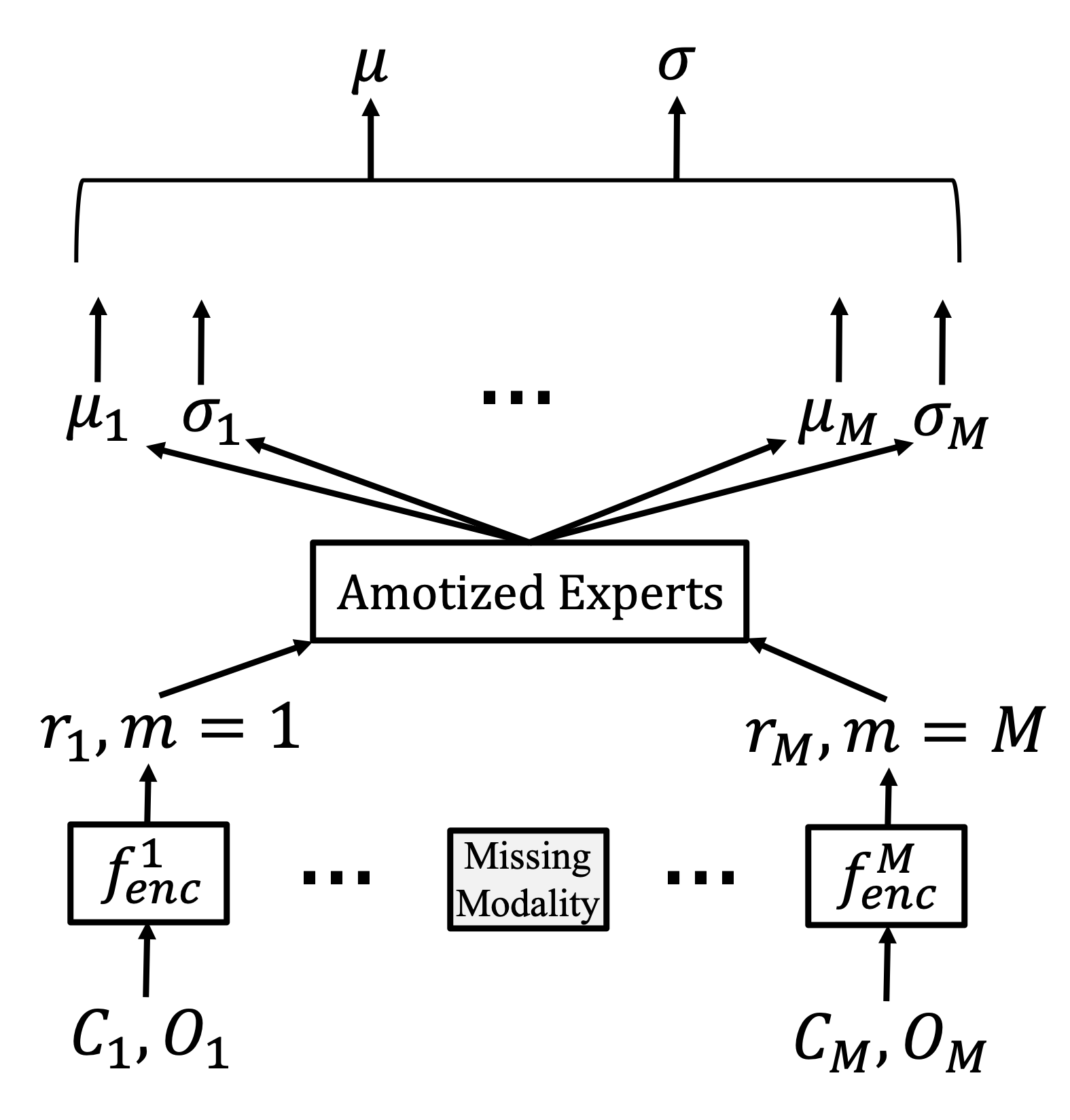}
    \hspace{-10mm}
    
    \vspace*{-0.15cm}
    
    \caption{
    Baseline model, PoE and APoE.
    {\small
    In the baseline model (left), a single inference network (denoted as \textit{Encoder}) get 
    an input as sum of all modality encoders's outputs.
    In PoE (middle), each of the experts contains an integrated network combining the modality encoder and a complex inference network like ConvDraw, resulting in $O(|M|)$ space cost of inference networks.
    In APoE (right), the modality encoding and the inference network are detached, and the inference networks are integrated into a single amortized expert inference network serving for all experts. Thus, the space cost of inference networks reduces to $O(1)$.
    }
    }
    \label{fig:baseline_poe_apoe}
\vspace{-0.3cm}
\end{figure}

\textbf{APoE Approximate Posterior.} The approximate posterior $Q_\phi(\bz | C,O)$ is implemented as follows. Following~\cite{WuG18/nips}, we first represent the true posterior as $P(\bz|C,O)=$
\eq{
    &\f{P(O,C|\bz)P(\bz)}{P(O,C)} = \f{P(\bz)}{P(C,O)}\prod_{m\in\cM_S}P(C_m,O_m|\bz)
    =\f{P(\bz)}{P(C,O)}\prod_{m\in\cM_S}\f{P(\bz|C_m,O_m)P(C_m,O_m)}{P(\bz)}.\nn
}
After ignoring the terms that are not a function of $\bz$, we obtain
$P(\bz|C,O) \propto \f{\prod_{m\in\cM_S}P(\bz|C_m,O_m)}{\pd{i}{|\cM_S|-1}P(\bz)}.$
Replacing the numerator terms with an approximation $P(\bz|C_m,O_m)\approx Q(\bz|C_m,O_m)P(\bz)^{\f{|\cM_S|-1}{|\cM_S|}}$, we can remove the priors in the denominator and obtain the following APoE approximate posterior:
\eq{P(\bz|C,O) \approx \prod_{m\in\cM_S}Q_\phi(\bz|C_m,O_m)\;.}
Although the above product is intractable in general, a closed form solution exists if each expert is a Gaussian~\citep{WuG18/nips}.~The mean $\mu$ and covariance $T$ of the APoE are, respectively, $\mu=(\sum_{m} \mu_m U_m)(\sum_{m} U_m)^{\inv}$ and $T=(\sum_{m} U_m)^{\inv}$, where $\mu_m$ and $U_m$ are the mean and the inverse of the covariance of each expert. The posterior APoE $Q_\phi(\bz|C_m,O_m)$ is implemented first by encoding $\br^m = f_\text{enc}^m(C_m,O_m)$ and then putting $\br^m$ and modality-id $m$ into the amortized expert $Q_\phi(\bz|\br^m,m)$, which is a ConvDraw module in our implementation. The amortized expert outputs $\mu_m$ and $U_m$ for $m\in\cM_S$ while sharing the variational parameter $\phi$ across the modality-experts. Fig. \ref{fig:baseline_poe_apoe} compares the inference network architectures of CGQN, PoE, and APoE.

\section{Related Works}\label{sc:related_works}
%

\textbf{Multimodal Generative Models.}
Multimodal data are associated with many interesting learning problems, \emph{e.g.} cross-modal inference, zero-shot learning or weakly-supervised learning. Regarding these, latent variable models have provided effective solutions: from a model with global latent variable shared among all modalities \citep{SuzukiNM16} to hierarchical latent structures \nolink{\citep{HsuG18}} and scalable inference networks with Product-of-Experts (PoE) \citep{Hinton02/neco, WuG18/nips, KurleGS18}.
In contrast to these works, the current study addresses two additional challenges. First, this work aims at achieving the any-modal to any-modal conditional inference regardless of modality configurations during training: it targets on generalization under distribution shifts at test time. On the other hand, the previous studies assume to have full modality configurations in training even when missing modality configuration at test time is addressed. %
Second, the proposed model considers each source of information to be rather partially observable, while each modality-specific data has been treated as fully observable. As a result, the modality-agnostic metamodal representation is inferred from modality-specific representations, each of which is integrated from a set of partially observable inputs.

\textbf{3D Representations and Rendering.}
Learning representations of 3D scenes or environments with partially observable inputs has been addressed by supervised learning \citep{choy20163d,wu2017marrnet,ShinFH18/cvpr, MeschederONNG18}, latent variable models \citep{Eslami1204/science18, RosenbaumBVRE2018, KumarERGVLS18}, and generative adversarial networks \citep{wu2016learning, RajeswarMGVNC19, NguyenPhuocLTRY19}.
The GAN-based approaches exploited domain-specific functions, \emph{e.g.} 3D representations, 3D-to-2D projection, and 3D rotations. Thus, it is hard to apply to non-visual modalities whose underlying transformations are unknown.
On the other hand, neural latent variable models for random processes \citep{Eslami1204/science18, RosenbaumBVRE2018, KumarERGVLS18, Garnelo2018a, Garnelo2018b, LeKGRST18/nipsworkshop, kim2018attentive/iclr}
has dealt with more generalized settings and studied on order-invariant inference.
However, these studies focus on single modality cases, so they are contrasted from our method, addressing a new problem setting where qualitatively different information sources are available for learning the scene representations.

\section{Experiment}

\begin{figure}[!tbt]
\centering
\hspace{-0.5cm}
\begin{subfigure}[t]{3.5cm}
    \includegraphics[width=3.50cm]{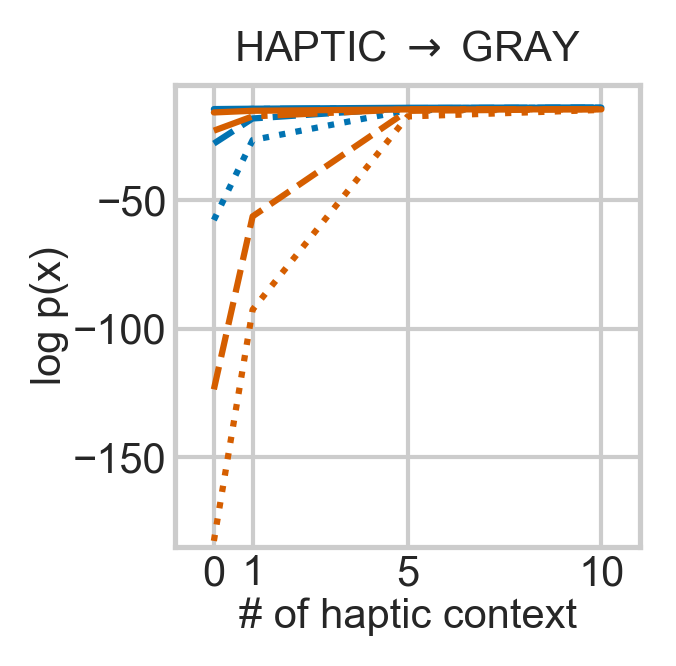} %
    \vspace{-1.6\baselineskip}
    \caption{}
\end{subfigure}
\hspace{-0.35cm}
\begin{subfigure}[t]{5.1cm}
    \includegraphics[width=5.1cm]{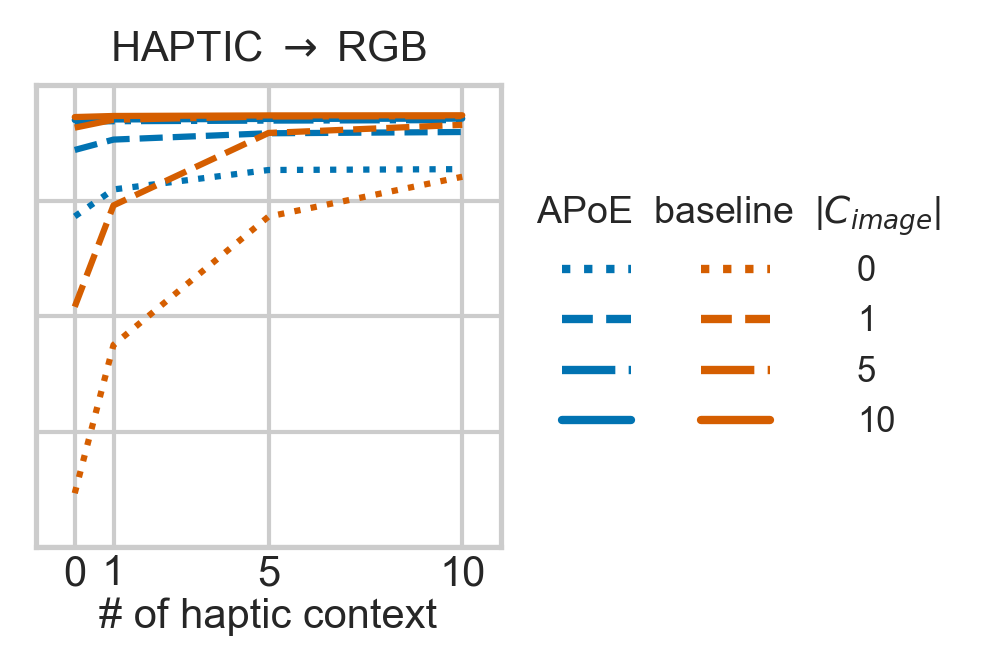} %
    \vspace{-1.6\baselineskip}
    \caption{$\quad\quad\quad\quad\quad\quad$}
\end{subfigure}
\hspace{-0.3cm}
\begin{subfigure}[t]{5.57cm}
    \includegraphics[width=5.57cm]{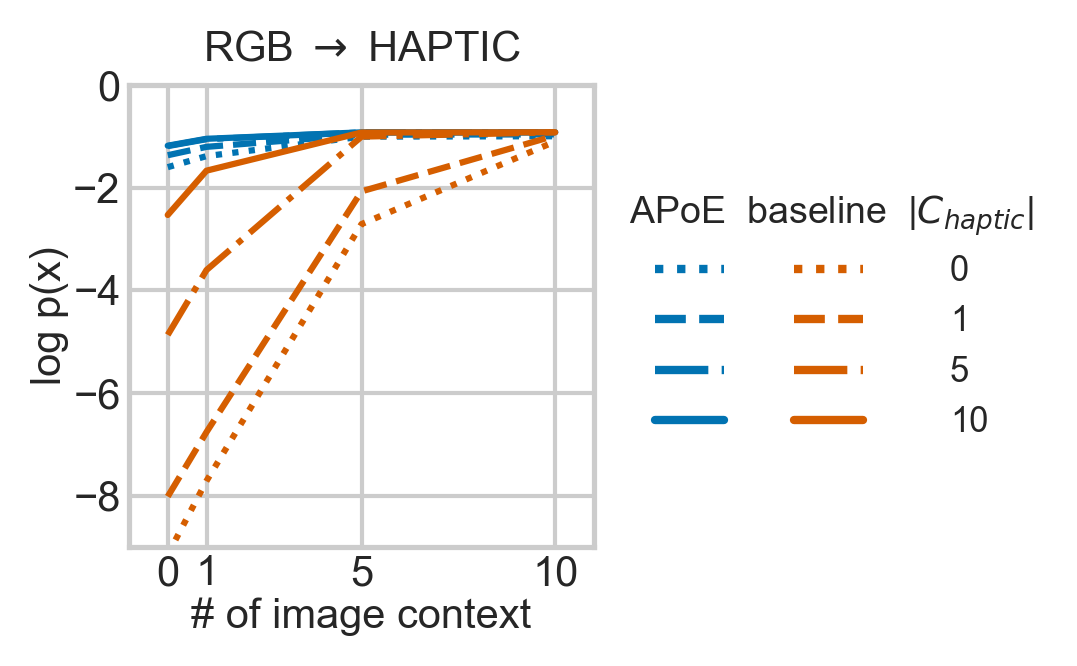} %
    \vspace{-1.6\baselineskip}
    \caption{$\quad\quad\quad\quad\quad$}
\end{subfigure}
\hspace{-0.35cm}
\vspace{-0.1cm}
\caption{
\small
Results on cross-modal density estimation.
\small
(a) log-likelihood of target images (gray) vs. the number of haptic observation.
(b) log-likelihood of target images (rgb) vs. the number of haptic observation.
(c) log-likelihood of target haptic values vs. the number of image observations.
The dotted lines show fully cross-modal inference where the context does not include any target modality.
For the inference with additional context from the target modality, the results are denoted as dashed, dashdot, and solid lines.
}
\label{fig:result-cross-modal-inference-log-likelihood}
\vspace{-0.5cm}
\end{figure}

The proposed model is evaluated with respect to the following criteria: (i) cross-modal density estimation in terms of log-likelihood, (ii) ability to perform cross-modal sample generation, (iii) quality of learned representation by applying it to a downstream classification task, (iv) robustness to the missing-modality problem, and (v) space and computational cost.

To evaluate our model we have developed an environment, the Multisensory Embodied 3D-Scene Environment (MESE).
MESE integrates MuJoCo~\citep{conf/iros/TodorovET12}, MuJoCo HAPTIX~\citep{KumarT15/humanoids}, and the OpenAI gym~\citep{BrockmanCPSSTZ16} for 3D scene understanding through multisensory interactions. In particular, from MuJoCo HAPTIX the Johns Hopkins Modular Prosthetic Limb (MPL) \citep{johannes2011overview} is used. The resulting MESE, equipped with vision and proprioceptive sensors, makes itself particularly suitable for tasks related to human-like embodied multisensory learning. In our experiments, the visual input is $64 \times 64$ RGB image and the haptic input is 132-dimension consisting of the hand pose and touch senses.
Our main task is similar to the Shepard-Metzler object experiments used in~\cite{Eslami1204/science18} but extends it with the MPL hand.

As a baseline model, we use a GQN variant~\citep{KumarERGVLS18} (discussed in Section~\ref{gen-process-prior-for-multisensory-context}). In this model, following GQN, the representations from different modalities are summed and then given to a ConvDraw network. We also provide a comparison to PoE version of the model in terms of computation speed and memory footprint. For more details on the experimental environments, implementations, and settings, refer to Appendix \ref{sec:appendix-experiments}.

\textbf{Cross-Modal Density Estimation.}~Our first evaluation is the cross-modal conditional density estimation.~For this, we estimate the conditional log-likelihood $\log P(X|V,C)$ for $\cM=\{\tt{RGB\textrm{-}image},\textrm{ }\tt{haptics}\}$, \emph{i.e.}~$|\cM|=2$.~During training, we use both modalities for each sampled scene and use 0 to 15 randomly sampled context query-sense pairs for each modality. At test time, we provide uni-modal context from one modality and generate the other.

Fig.~\ref{fig:result-cross-modal-inference-log-likelihood} shows results on 3 different experiments: (a) HAPTIC$\ra$GRAY, (b) HAPTIC$\ra$RGB and (c) RGB$\ra$HAPTIC. Note that we include HAPTIC$\ra$GRAY -- although GRAY images are not used during training -- to analyze the effect of color in haptic-to-image generation. The APoE and the baseline are plotted in blue and orange, respectively. 
In all cases our model (blue) outperforms the baseline (orange). This gap is even larger when the model is provided limited amount of context information, suggesting that the baseline requires more context to improve the representation. Specifically, in the fully cross modal setting where the context does not include any target modality (the dotted lines), the gap is largest. We believe that our model can better leverage modal-invariant representations from one modality to another.
Also, when we provide additional context from the target modality (dashed, dashdot, solid lines), we still see that our model outperforms the baseline. This implies that our models can successfully incorporate information from different modalities without interfering each other. Furthermore, from Fig.~\ref{fig:result-cross-modal-inference-log-likelihood}(a) and (b), we observe that haptic information captures only shapes: the prediction in RGB has lower likelihood without any image in the context. However, for the GRAY image in (a), the likelihood approaches near the upper bound.

\textbf{Cross-Modal Generation.} We now qualitatively evaluate the ability for cross-generation.
Fig.~\ref{fig:result-cross-modal-generation} shows samples of our cross-modal generation for various query viewpoints. Here, we condition the model on 15 haptic context signal but provide only a single image.
We note that the single image provides limited color information about the object, namely, red and cyan are part of the object and almost no information about the shape.
We can see that the model is able to almost perfectly infer the shape of the object. However, it fails to predict the correct colors (Fig.~\ref{fig:result-cross-modal-generation}(c)) which is expected due to the limited visual information provided. Interestingly, the object part for which the context image provides color information has correct colors, while other parts have random colors for different samples, showing that the model captures the uncertainty in $\bz$.
Additional results provided in the Appendix~\ref{sec:appendix-cross-modal-generation} suggest further that: (i) our model gradually aggregates numerous evidences to improve predictions (Fig.~\ref{fig:result-reducing-uncertainity-with-aggregration}) and (ii) our model successfully integrates distinctive multisensory information in their inference (Fig.~\ref{fig:result-cross-modal-generation-extended}).

\begingroup
\setlength{\intextsep}{0pt}

\begin{wrapfigure}{R}{0.40\textwidth}
\centering
\includegraphics[width=0.40\textwidth]{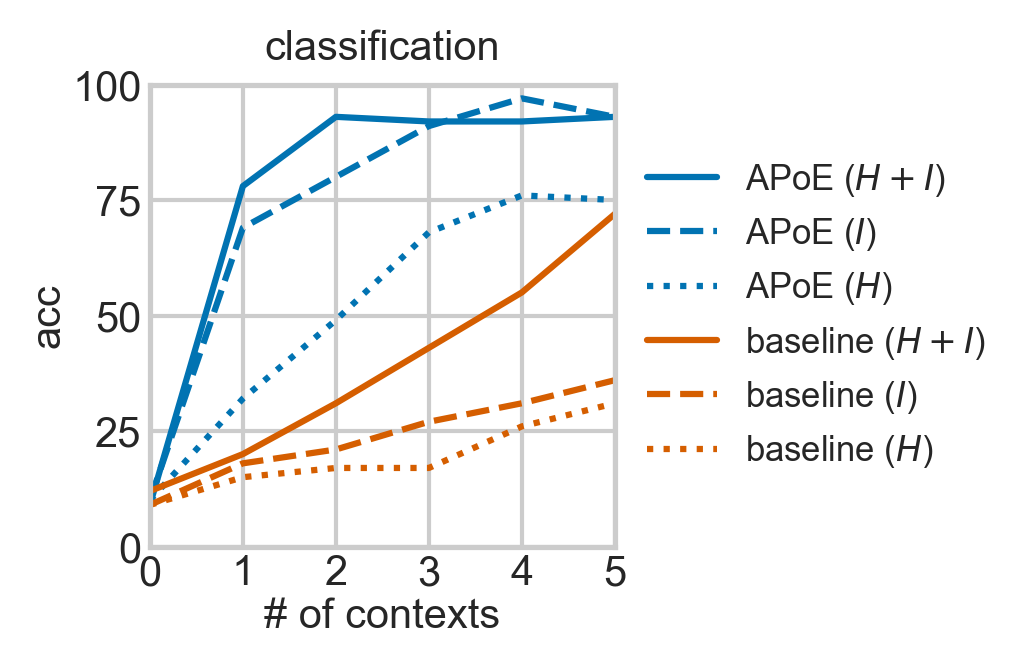} %
\vspace{-1.6\baselineskip}
\caption{
Classification result.
}
\label{fig:result-classification}
\end{wrapfigure}
\textbf{Classification.}~To further evaluate the quality of the modality-invariant scene representations, we test on a downstream classification task.
We randomly sample 10 scenes and from each scene we prepare a held-out query-sense pairs to use as the input to the classifier. The models are then asked to classify which scene (1 out of 10) a given query-sense pair belongs to. We use Eq. \eqref{eq:classification} for this classification.
To see how the provided multi-modal context contributes to obtaining useful representation for this task, we test the following three context configurations: (i) image-query pairs only ($I$), (ii) haptic-query pairs only ($H$), and (iii) all sensory contexts ($H+I$).

\endgroup

In Fig.~\ref{fig:result-classification}, both models use contexts to classify scenes and their performance improves as the number of contexts increases. APoE outperforms the baseline in the classification accuracy, while both methods have similar ELBO (see Fig. \ref{fig:result-elbo-classification-m2}). This suggests that the representation of our model tends to be more discriminative than that of the baseline.
In APoE, the results with individual modality ($I$ or $H$) are close to the one with all modalities ($H+I$). 
The drop in performance with only haptic-query pairs ($H$) is due to the fact that certain samples might have same shape, but different colors.
On the other hand, the baseline shows worse performance when inferring modality-invariant representation with single sensory modality, especially for images.~This demonstrates that the APoE model helps learning better representations for both modality-specific ($I$ and $H$) and modality-invariant tasks ($H+I$).

\begin{figure}[!tb]
\centering
\begin{subfigure}[t]{3.10cm}
    \includegraphics[width=3.10cm]{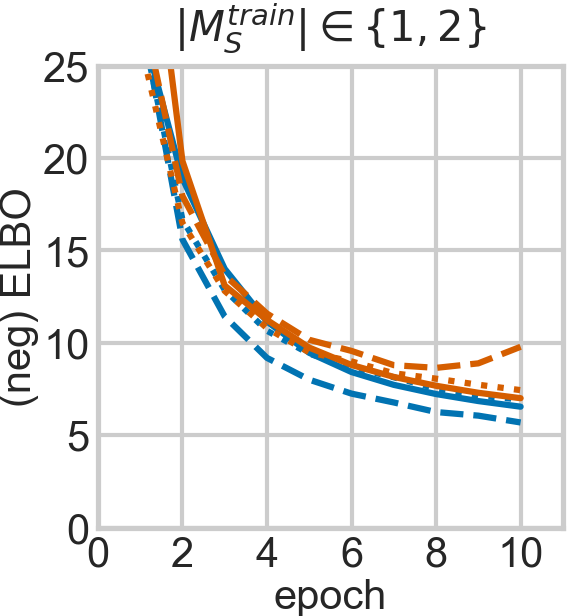}
    \vspace{-1.2\baselineskip}
    \caption{}
\end{subfigure}
\begin{subfigure}[t]{2.63cm}
    \includegraphics[width=2.63cm]{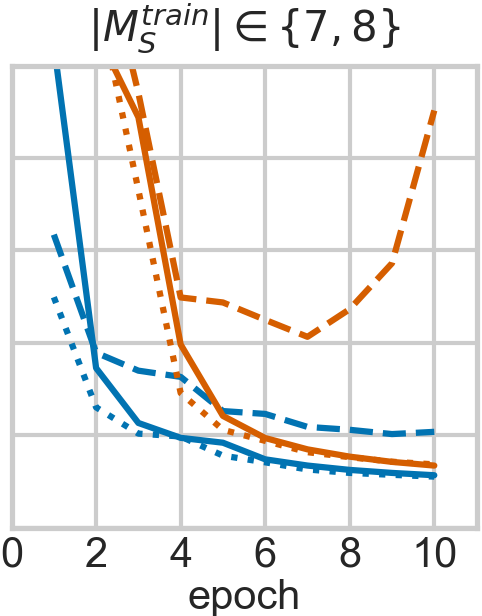}
    \vspace{-1.2\baselineskip}
    \caption{}
\end{subfigure}
\begin{subfigure}[t]{2.63cm}
    \includegraphics[width=2.63cm]{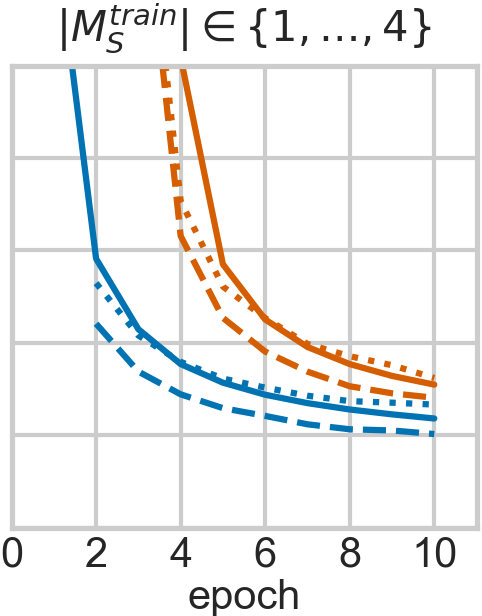} %
    \vspace{-1.2\baselineskip}
    \caption{}
\end{subfigure}
\begin{subfigure}[t]{5.06cm}
    \includegraphics[width=5.06cm]{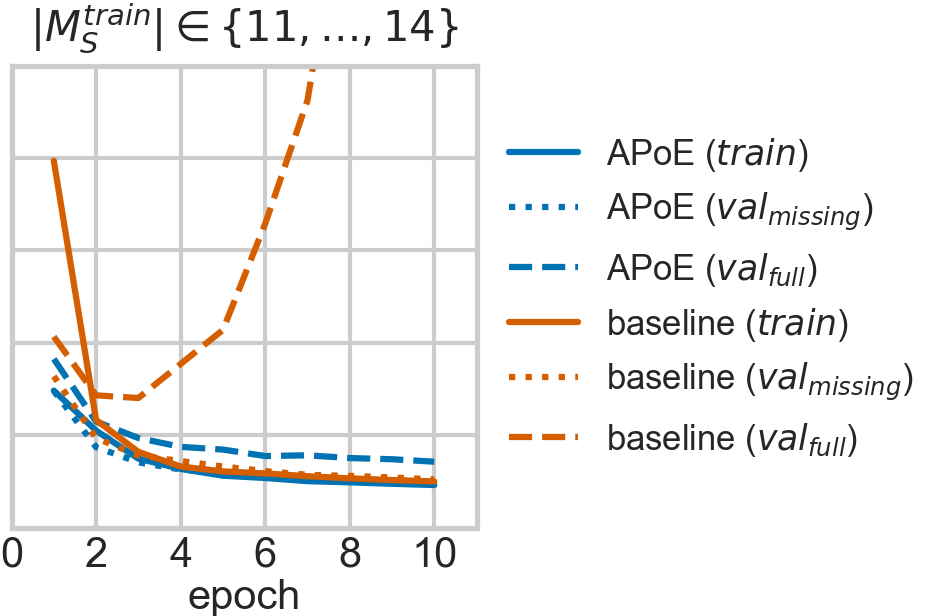} %
    \vspace{-1.2\baselineskip}
    \caption{$\quad\quad\quad\quad\quad\quad\quad$}
\end{subfigure}
\vspace{-0.2cm}
\caption{
Results of missing-modality experiments for (a,b) $|\cM|=8$, and (c,d) $14$ environments.
{\small
During training ($\tt{train}$), limited combinations of all possible modalities are presented to the model.
The size of exposed multimodal senses per scene is denoted as $|\cM_{S}^{\tt{train}}|$.
For validation dataset, the models are evaluated with the same limited combinations as done in training (${\tt{val}}_{\tt{missing}}$), as well as all combinations (${\tt{val}}_{\tt{full}}$).
}
}
\label{fig:result-extrapolation-interpolation}
\vspace{-0.5cm}
\end{figure}

\textbf{Missing-modality Problem.}
In practical scenarios, since it is difficult to assume that we always have access to all modalities, it is important to make the model learn when some modalities are missing. 
Here, we evaluate this robustness by providing unseen combinations of modalities at test time. 
This is done by limiting the set of modality combinations observed during training. That is, we provide only a subset of modality combinations for each scene $S$, \emph{i.e}, $\cM_{S}^{\tt{train}} \subset \cM$. At test time, the model is evaluated on every combinations of all modalities $\cM$ thus including the settings not observed during training.
As an example, for total 8 modalities $\cM = \{\tt{left, right}$\footnote{left and right half of an image}$\} \times \{\tt{R, G, B}\} \times \{\tt{haptics}_1\} \times  \{\tt{haptics}_2$\}, we use $|\cM_{S}^{\tt{train}}| \in \{1, 2\}$ to indicate that each scene in training data contains only one or two modalities. Fig.~\ref{fig:result-extrapolation-interpolation}(a) and (b) show results with $|\cM|=8$ while (c) and (d) with $|\cM|=14$.

Fig.~\ref{fig:result-extrapolation-interpolation} (a) and (c) are results when a much more restricted number of modalities are available during training: 2 out of 8 and 4 out of 14, respectively.
At test time, however, all combinations of modalities are used. We denote the performance on the full configurations by ${\tt{val}}_{\tt{full}}$ and on the limited modality configurations used during training by ${\tt{val}}_{\tt{missing}}$. Fig.~\ref{fig:result-extrapolation-interpolation} (b) and (d) show the opposite setting where, during training, a large number of modalities (e.g., 7$\sim$8 modalities) are always provided together for each scene. Thus, the scenes have not trained on small modalities such as only one or two modalities but we tested on this configurations at test time to see its ability to learn to perform the generalized cross-modal generation. For more results, see Appendix \ref{sec:appendix-incomplete-multisensory-data}.

Overall, for all cases our model shows good test time performance on the unseen context modality configurations whereas the baseline model mostly overfits (except (c)) severely or converges slowly. This is because, in the baseline model, the sum representation on the unseen context configuration is likely to be also unseen at test time and thus overfit. In contrast, our model as a PoE is robust to this problem as all experts agree to make a similar representation. The baseline results for case (c) seem less prone to this problem but converged much slowly. As it converges slowly, we believe that it might still overfit in the end with a longer run.

\textbf{Space and Time Complexity.}
The expert amortizer of APoE significantly reduces the inherent space problem of PoE while it still requires separate modality encoders. Specifically, in our experiments, for the $\cM=5$ case, PoE requires 53M parameters while APoE uses 29M. For $\cM=14$, PoE uses 131M parameters while APoE used only 51M. We also observed a reduction in computation time by using APoE. For $\cM=5$ model, one iteration of PoE takes, in average, 790 ms while APoE takes 679 ms. This gap becomes more significant for $\cM=14$ where PoE takes 2059 ms while APoE takes 1189 ms. This is partly due to the number of parameters. Moreover, unlike PoE, APoE can parallelize its encoder computation via convolution. For more results, see Table~\ref{tab:computation-costs} in Appendix.

\section{Conclusion}
We propose the \emph{Generative Multisensory Network} (GMN) for understanding 3D scenes via 
modality-invariant representation learning. In GMN, we introduce the \emph{Amortized Product-of-Experts} (APoE) in order to deal with the problem of missing-modalities while resolving the space complexity problem of standard Product-of-Experts. In experiments on 3D scenes with blocks of different shapes and a human-like hand, we show that GMN can generate any modality from any context configurations.
We also show that the model with APoE learns better modality-agnostic representations, as well as modality-specific ones.
To the best of our knowledge this is the first exploration of multisensory representation learning with vision and haptics for generating 3D objects. Furthermore, we have developed a novel multisensory simulation environment, called the Multisensory Embodied 3D-Scene Environment (MESE), that is critical to performing these experiments.

\subsubsection*{Acknowledgments}
JL would like to thank Chin-Wei Huang, Shawn Tan, Tatjana Chavdarova, Arantxa Casanova, Ankesh Anand, and Evan Racah for helpful comments and advice. SA thanks Kakao Brain, the Center for Super Intelligence (CSI), and Element AI for their support. CP also thanks NSERC and PROMPT.

\bibliographystyle{iclr}
\bibliography{refs}

\begin{thebibliography}{46}
\providecommand{\natexlab}[1]{#1}
\providecommand{\url}[1]{\texttt{#1}}
\expandafter\ifx\csname urlstyle\endcsname\relax
  \providecommand{\doi}[1]{doi: #1}\else
  \providecommand{\doi}{doi: \begingroup \urlstyle{rm}\Url}\fi

\bibitem[Amos et~al.(2018)Amos, Dinh, Cabi, Rothörl, Muldal, Erez, Tassa,
  de~Freitas, and Denil]{amos2018learning}
Brandon Amos, Laurent Dinh, Serkan Cabi, Thomas Rothörl, Alistair Muldal, Tom
  Erez, Yuval Tassa, Nando de~Freitas, and Misha Denil.
\newblock Learning awareness models.
\newblock In \emph{ICLR}, 2018.

\bibitem[Barsalou(2008)]{barsalou2008grounded}
Lawrence~W Barsalou.
\newblock Grounded cognition.
\newblock \emph{Annu. Rev. Psychol.}, 2008.

\bibitem[Brockman et~al.(2016)Brockman, Cheung, Pettersson, Schneider,
  Schulman, Tang, and Zaremba]{BrockmanCPSSTZ16}
Greg Brockman, Vicki Cheung, Ludwig Pettersson, Jonas Schneider, John Schulman,
  Jie Tang, and Wojciech Zaremba.
\newblock Openai gym.
\newblock \emph{arXiv preprint arXiv:1606.01540}, 2016.

\bibitem[Burda et~al.(2016)Burda, Grosse, and Salakhutdinov]{BurdaGS15/iclr}
Yuri Burda, Roger~B. Grosse, and Ruslan Salakhutdinov.
\newblock Importance weighted autoencoders.
\newblock In \emph{ICLR}, 2016.

\bibitem[Chetlur et~al.(2014)Chetlur, Woolley, Vandermersch, Cohen, Tran,
  Catanzaro, and Shelhamer]{chetlur2014cudnn}
Sharan Chetlur, Cliff Woolley, Philippe Vandermersch, Jonathan Cohen, John
  Tran, Bryan Catanzaro, and Evan Shelhamer.
\newblock cudnn: Efficient primitives for deep learning.
\newblock \emph{arXiv preprint arXiv:1410.0759}, 2014.

\bibitem[Choy et~al.(2016)Choy, Xu, Gwak, Chen, and Savarese]{choy20163d}
Christopher~B Choy, Danfei Xu, JunYoung Gwak, Kevin Chen, and Silvio Savarese.
\newblock 3d-r2n2: A unified approach for single and multi-view 3d object
  reconstruction.
\newblock In \emph{ECCV}, 2016.

\bibitem[Deneve \& Pouget(2004)Deneve and Pouget]{deneve2004bayesian}
Sophie Deneve and Alexandre Pouget.
\newblock Bayesian multisensory integration and cross-modal spatial links.
\newblock \emph{Journal of Physiology-Paris}, 98\penalty0 (1-3):\penalty0
  249--258, 2004.

\bibitem[Eslami et~al.(2018)Eslami, Jimenez~Rezende, Besse, Viola, Morcos,
  Garnelo, Ruderman, Rusu, Danihelka, Gregor, Reichert, Buesing, Weber,
  Vinyals, Rosenbaum, Rabinowitz, King, Hillier, Botvinick, Wierstra,
  Kavukcuoglu, and Hassabis]{Eslami1204/science18}
S.~M.~Ali Eslami, Danilo Jimenez~Rezende, Frederic Besse, Fabio Viola, Ari~S.
  Morcos, Marta Garnelo, Avraham Ruderman, Andrei~A. Rusu, Ivo Danihelka, Karol
  Gregor, David~P. Reichert, Lars Buesing, Theophane Weber, Oriol Vinyals, Dan
  Rosenbaum, Neil Rabinowitz, Helen King, Chloe Hillier, Matt Botvinick, Daan
  Wierstra, Koray Kavukcuoglu, and Demis Hassabis.
\newblock Neural scene representation and rendering.
\newblock \emph{Science}, 2018.

\bibitem[Garnelo et~al.(2018{\natexlab{a}})Garnelo, Rosenbaum, Maddison,
  Ramalho, Saxton, Shanahan, Teh, Rezende, and Eslami]{Garnelo2018a}
Marta Garnelo, Dan Rosenbaum, Chris~J. Maddison, Tiago Ramalho, David Saxton,
  Murray Shanahan, Yee~Whye Teh, Danilo~J. Rezende, and S.~M.~Ali Eslami.
\newblock Conditional neural processes.
\newblock \emph{arXiv preprint arXiv:1807.01613}, 2018{\natexlab{a}}.

\bibitem[Garnelo et~al.(2018{\natexlab{b}})Garnelo, Schwarz, Rosenbaum, Viola,
  Rezende, Eslami, and Teh]{Garnelo2018b}
Marta Garnelo, Jonathan Schwarz, Dan Rosenbaum, Fabio Viola, Danilo~J. Rezende,
  S.~M.~Ali Eslami, and Yee~Whye Teh.
\newblock Neural processes.
\newblock \emph{arXiv preprint arXiv:1807.01622}, 2018{\natexlab{b}}.

\bibitem[Gregor et~al.(2016)Gregor, Besse, Rezende, Danihelka, and
  Wierstra]{GregorBRDW16/nips}
Karol Gregor, Frederic Besse, Danilo~Jimenez Rezende, Ivo Danihelka, and Daan
  Wierstra.
\newblock Towards conceptual compression.
\newblock In \emph{NIPS}, 2016.

\bibitem[Higgins et~al.(2017)Higgins, Matthey, Pal, Burgess, Glorot, Botvinick,
  Mohamed, and Lerchner]{higgins2017beta}
Irina Higgins, Loic Matthey, Arka Pal, Christopher Burgess, Xavier Glorot,
  Matthew Botvinick, Shakir Mohamed, and Alexander Lerchner.
\newblock beta-vae: Learning basic visual concepts with a constrained
  variational framework.
\newblock In \emph{ICLR}, 2017.

\bibitem[Hinton(2002)]{Hinton02/neco}
Geoffrey~E. Hinton.
\newblock Training products of experts by minimizing contrastive divergence.
\newblock \emph{Neural Computation}, 2002.

\bibitem[Hsu \& Glass(2018)Hsu and Glass]{HsuG18}
Wei{-}Ning Hsu and James~R. Glass.
\newblock Disentangling by partitioning: {A} representation learning framework
  for multimodal sensory data.
\newblock \emph{arXiv preprint arXiv:1805.11264}, 2018.

\bibitem[Johannes et~al.(2011)Johannes, Bigelow, Burck, Harshbarger, Kozlowski,
  and Van~Doren]{johannes2011overview}
Matthew~S Johannes, John~D Bigelow, James~M Burck, Stuart~D Harshbarger,
  Matthew~V Kozlowski, and Thomas Van~Doren.
\newblock An overview of the developmental process for the modular prosthetic
  limb.
\newblock \emph{Johns Hopkins APL Technical Digest}, 2011.

\bibitem[Kim et~al.(2019)Kim, Mnih, Schwarz, Garnelo, Eslami, Rosenbaum,
  Vinyals, and Teh]{kim2018attentive/iclr}
Hyunjik Kim, Andriy Mnih, Jonathan Schwarz, Marta Garnelo, Ali Eslami, Dan
  Rosenbaum, Oriol Vinyals, and Yee~Whye Teh.
\newblock Attentive neural processes.
\newblock In \emph{ICLR}, 2019.

\bibitem[Kingma \& Ba(2014)Kingma and Ba]{KingmaB14}
Diederik~P. Kingma and Jimmy Ba.
\newblock Adam: {A} method for stochastic optimization.
\newblock \emph{arXiv preprint arXiv:1412.6980}, 2014.

\bibitem[Kingma \& Welling(2013)Kingma and Welling]{kingma2013auto}
Diederik~P Kingma and Max Welling.
\newblock Auto-encoding variational bayes.
\newblock \emph{arXiv preprint arXiv:1312.6114}, 2013.

\bibitem[Konečný et~al.(2016)Konečný, McMahan, Yu, Richtarik, Suresh, and
  Bacon]{fedlearn16}
Jakub Konečný, H.~Brendan McMahan, Felix~X. Yu, Peter Richtarik,
  Ananda~Theertha Suresh, and Dave Bacon.
\newblock Federated learning: Strategies for improving communication
  efficiency.
\newblock In \emph{NIPS Workshop on Private Multi-Party Machine Learning},
  2016.
\newblock URL \url{https://arxiv.org/abs/1610.05492}.

\bibitem[Kumar et~al.(2018)Kumar, Eslami, Rezende, Garnelo, Viola, Lockhart,
  and Shanahan]{KumarERGVLS18}
Ananya Kumar, S.~M.~Ali Eslami, Danilo~J. Rezende, Marta Garnelo, Fabio Viola,
  Edward Lockhart, and Murray Shanahan.
\newblock Consistent generative query networks.
\newblock \emph{arXiv preprint arXiv:1807.02033}, 2018.

\bibitem[Kumar \& Todorov(2015)Kumar and Todorov]{KumarT15/humanoids}
Vikash Kumar and Emanuel Todorov.
\newblock Mujoco {HAPTIX:} {A} virtual reality system for hand manipulation.
\newblock In \emph{International Conference on Humanoid Robots, Humanoids},
  2015.

\bibitem[Kurle et~al.(2018)Kurle, G{\"{u}}nnemann, and van~der
  Smagt]{KurleGS18}
Richard Kurle, Stephan G{\"{u}}nnemann, and Patrick van~der Smagt.
\newblock Multi-source neural variational inference.
\newblock \emph{arXiv preprint arXiv:1811.04451}, 2018.

\bibitem[Lake et~al.(2015)Lake, Salakhutdinov, and
  Tenenbaum]{lake2015human/science}
Brenden~M Lake, Ruslan Salakhutdinov, and Joshua~B Tenenbaum.
\newblock Human-level concept learning through probabilistic program induction.
\newblock \emph{Science}, 2015.

\bibitem[Le et~al.(2018)Le, Kim, Garnelo, Rosenbaum, Schwarz, and
  Teh]{LeKGRST18/nipsworkshop}
Tuan~Anh Le, Hyunjik Kim, Marta Garnelo, Dan Rosenbaum, Jonathan Schwarz, and
  Yee~Whye Teh.
\newblock Empirical evaluation of neural process objectives.
\newblock In \emph{NeurIPS Bayesian Workshop}, 2018.

\bibitem[Mescheder et~al.(2018)Mescheder, Oechsle, Niemeyer, Nowozin, and
  Geiger]{MeschederONNG18}
Lars~M. Mescheder, Michael Oechsle, Michael Niemeyer, Sebastian Nowozin, and
  Andreas Geiger.
\newblock Occupancy networks: Learning 3d reconstruction in function space.
\newblock \emph{arXiv preprint arXiv:1812.03828}, 2018.

\bibitem[Murray \& Wallace(2011)Murray and Wallace]{murray2011neural}
Micah~M Murray and Mark~T Wallace.
\newblock \emph{The neural bases of multisensory processes}.
\newblock CRC Press, 2011.

\bibitem[Nguyen{-}Phuoc et~al.(2019)Nguyen{-}Phuoc, Li, Theis, Richardt, and
  Yang]{NguyenPhuocLTRY19}
Thu Nguyen{-}Phuoc, Chuan Li, Lucas Theis, Christian Richardt, and Yong{-}Liang
  Yang.
\newblock Hologan: Unsupervised learning of 3d representations from natural
  images.
\newblock \emph{arXiv preprint arXiv:1904.01326}, 2019.

\bibitem[Nickolls et~al.(2008)Nickolls, Buck, Garland, and
  Skadron]{nickolls2008cuda}
John Nickolls, Ian Buck, Michael Garland, and Kevin Skadron.
\newblock Scalable parallel programming with cuda.
\newblock \emph{Queue}, 6\penalty0 (2):\penalty0 40--53, March 2008.
\newblock ISSN 1542-7730.

\bibitem[Pascual-Leone \& Hamilton(2001)Pascual-Leone and
  Hamilton]{pascual2001metamodal}
Alvaro Pascual-Leone and Roy Hamilton.
\newblock The metamodal organization of the brain.
\newblock In \emph{Progress in brain research}, volume 134, pp.\  427--445.
  Elsevier, 2001.

\bibitem[Paszke et~al.(2017)Paszke, Gross, Chintala, Chanan, Yang, DeVito, Lin,
  Desmaison, Antiga, and Lerer]{paszke2017pytorch}
Adam Paszke, Sam Gross, Soumith Chintala, Gregory Chanan, Edward Yang, Zachary
  DeVito, Zeming Lin, Alban Desmaison, Luca Antiga, and Adam Lerer.
\newblock Automatic differentiation in pytorch.
\newblock In \emph{NIPS-W}, 2017.

\bibitem[Quiroga(2012)]{quiroga2012concept}
Rodrigo~Quian Quiroga.
\newblock Concept cells: the building blocks of declarative memory functions.
\newblock \emph{Nature Reviews Neuroscience}, 2012.

\bibitem[Rajeswar et~al.(2019)Rajeswar, Mannan, Golemo, Vazquez,
  Nowrouzezahrai, and Courville]{RajeswarMGVNC19}
Sai Rajeswar, Fahim Mannan, Florian Golemo, David Vazquez, Derek
  Nowrouzezahrai, and Aaron Courville.
\newblock Pix2scene: Learning implicit 3d representations from images.
\newblock \emph{preprint https://openreview.net/forum?id=BJeem3C9F7}, 2019.

\bibitem[Ramachandran \& Hirstein(1998)Ramachandran and
  Hirstein]{ramachandran1998perception}
Vilayanur~S Ramachandran and William Hirstein.
\newblock The perception of phantom limbs. the do hebb lecture.
\newblock \emph{Brain: a journal of neurology}, 121\penalty0 (9):\penalty0
  1603--1630, 1998.

\bibitem[Rezende et~al.(2014)Rezende, Mohamed, and
  Wierstra]{rezende2014stochastic}
Danilo~Jimenez Rezende, Shakir Mohamed, and Daan Wierstra.
\newblock Stochastic backpropagation and approximate inference in deep
  generative models.
\newblock \emph{arXiv preprint arXiv:1401.4082}, 2014.

\bibitem[Rohe \& Noppeney(2016)Rohe and Noppeney]{rohe2016distinct}
Tim Rohe and Uta Noppeney.
\newblock Distinct computational principles govern multisensory integration in
  primary sensory and association cortices.
\newblock \emph{Current Biology}, 26\penalty0 (4):\penalty0 509--514, 2016.

\bibitem[Rosenbaum et~al.(2018)Rosenbaum, Besse, Viola, Rezende, and
  Eslami]{RosenbaumBVRE2018}
Dan Rosenbaum, Frederic Besse, Fabio Viola, Danilo~J. Rezende, and S.~M.~Ali
  Eslami.
\newblock Learning models for visual 3d localization with implicit mapping.
\newblock \emph{arXiv preprint arXiv:1807.03149}, 2018.

\bibitem[Saha \& Dasgupta(2018)Saha and Dasgupta]{saha2018comprehensive}
Olimpiya Saha and Prithviraj Dasgupta.
\newblock A comprehensive survey of recent trends in cloud robotics
  architectures and applications.
\newblock \emph{Robotics}, 7\penalty0 (3):\penalty0 47, 2018.

\bibitem[Shams \& Seitz(2008)Shams and Seitz]{shams2008benefits}
Ladan Shams and Aaron~R Seitz.
\newblock Benefits of multisensory learning.
\newblock \emph{Trends in cognitive sciences}, 12\penalty0 (11):\penalty0
  411--417, 2008.

\bibitem[Shin et~al.(2018)Shin, Fowlkes, and Hoiem]{ShinFH18/cvpr}
Daeyun Shin, Charless~C. Fowlkes, and Derek Hoiem.
\newblock Pixels, voxels, and views: {A} study of shape representations for
  single view 3d object shape prediction.
\newblock In \emph{CVPR}, 2018.

\bibitem[Silver et~al.(2016)Silver, Huang, Maddison, Guez, Sifre, Van
  Den~Driessche, Schrittwieser, Antonoglou, Panneershelvam, Lanctot,
  et~al.]{silver2016mastering}
David Silver, Aja Huang, Chris~J Maddison, Arthur Guez, Laurent Sifre, George
  Van Den~Driessche, Julian Schrittwieser, Ioannis Antonoglou, Veda
  Panneershelvam, Marc Lanctot, et~al.
\newblock Mastering the game of go with deep neural networks and tree search.
\newblock \emph{nature}, 2016.

\bibitem[Suzuki et~al.(2016)Suzuki, Nakayama, and Matsuo]{SuzukiNM16}
Masahiro Suzuki, Kotaro Nakayama, and Yutaka Matsuo.
\newblock Joint multimodal learning with deep generative models.
\newblock \emph{arXiv preprint arXiv:1611.01891}, 2016.

\bibitem[Todorov et~al.(2012)Todorov, Erez, and Tassa]{conf/iros/TodorovET12}
Emanuel Todorov, Tom Erez, and Yuval Tassa.
\newblock Mujoco: A physics engine for model-based control.
\newblock In \emph{IROS}, 2012.

\bibitem[Wu et~al.(2016)Wu, Zhang, Xue, Freeman, and Tenenbaum]{wu2016learning}
Jiajun Wu, Chengkai Zhang, Tianfan Xue, Bill Freeman, and Josh Tenenbaum.
\newblock Learning a probabilistic latent space of object shapes via 3d
  generative-adversarial modeling.
\newblock In \emph{NIPS}, 2016.

\bibitem[Wu et~al.(2017)Wu, Wang, Xue, Sun, Freeman, and
  Tenenbaum]{wu2017marrnet}
Jiajun Wu, Yifan Wang, Tianfan Xue, Xingyuan Sun, Bill Freeman, and Josh
  Tenenbaum.
\newblock Marrnet: 3d shape reconstruction via 2.5 d sketches.
\newblock In \emph{NIPS}, 2017.

\bibitem[Wu \& Goodman(2018)Wu and Goodman]{WuG18/nips}
Mike Wu and Noah Goodman.
\newblock Multimodal generative models for scalable weakly-supervised learning.
\newblock In \emph{NeurIPS}, 2018.

\bibitem[Yildirim(2014)]{yildirim2014perception}
Ilker Yildirim.
\newblock \emph{From perception to conception: learning multisensory
  representations}.
\newblock University of Rochester, 2014.

\end{thebibliography}

\clearpage
\appendix
\renewcommand{\thefigure}{S\arabic{figure}}
\setcounter{figure}{0}

\section{Experiments}
\label{sec:appendix-experiments}
We start from describing the Multisensory Embodied 3D-Scene Environment (MESE) environment and simulated datasets used in our experiments. We continue by explaining training settings.

\begin{figure}[!h]
\centering
\includegraphics[width=7cm]{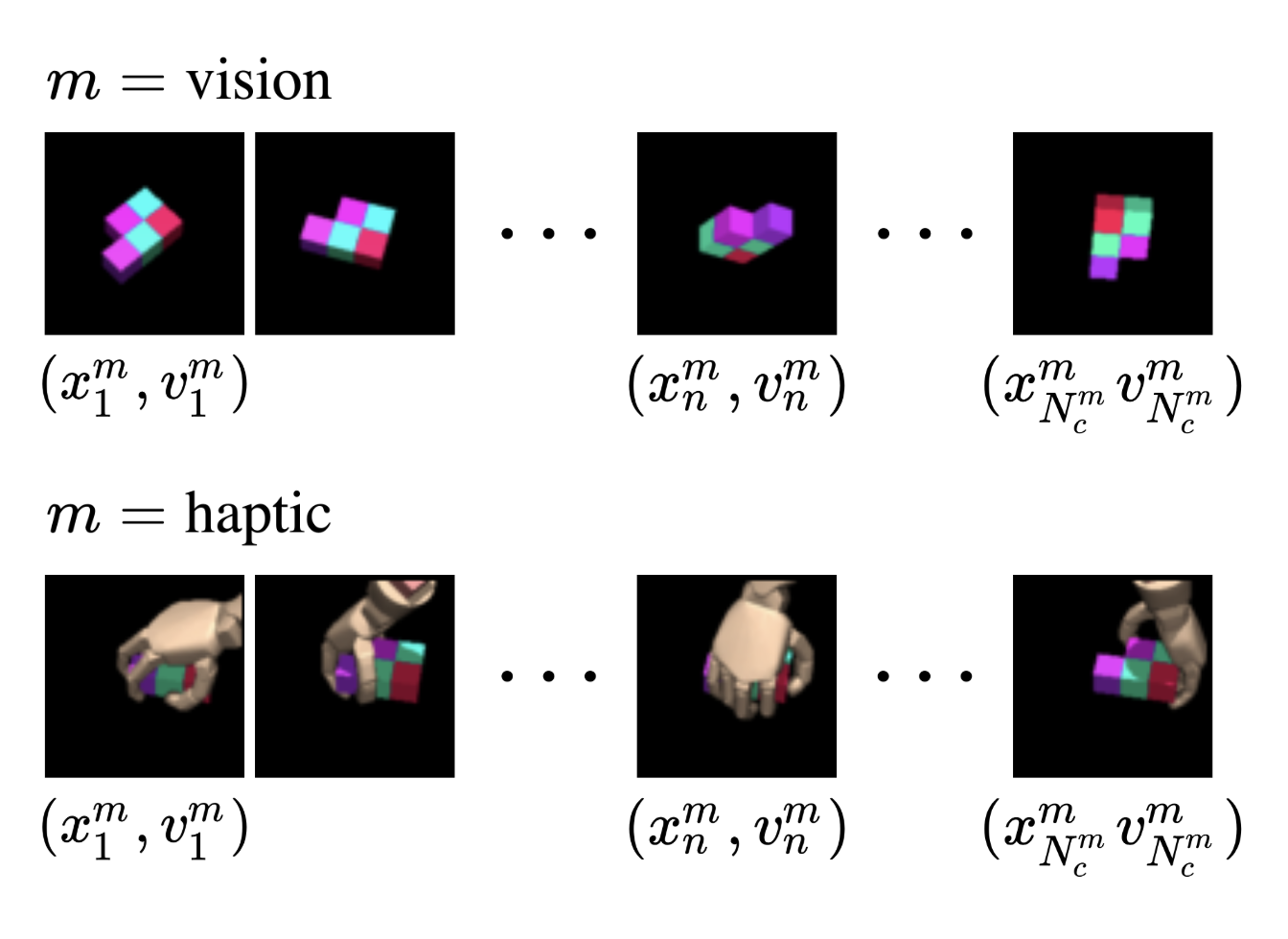}
\vspace{-0.3cm}
\caption{
Example multisensory scene (with single object) in MESE.
The scene includes a set of visual and haptic observations, each of which is partially observable.
}
\label{fig:multisensory-scene}
\end{figure}

\subsection{Multisensory Embodied 3D-Scene Environment (MESE)}
\label{sec:appendix-mese}
Targeting on a development environment for 3D scene understanding through interaction, we build a multisensory 3D scene environment, equipped with visual and proprioceptive (haptic) sensors, called Multisensory Embodied 3D-Scene Environment (MESE).
MESE is similar to Shepard-Metzler object experiments used in~\cite{Eslami1204/science18}, but extends it with a MPL hand model of MuJoCo HAPTIX \citep{KumarT15/humanoids}.
The environment uses MuJoCo~\citep{conf/iros/TodorovET12} and the OpenAI gym~\citep{BrockmanCPSSTZ16}.

\textbf{Scene.}
Adopted from \cite{Eslami1204/science18}, MESE generates single Shepard-Metzler object with an arbitrary number of blocks per episode.
Each block of the object is randomly colored in HSV scheme.
More precisely, hue and saturation are randomly selected within fixed ranges: hue is sampled from (0, 1) and saturation is sampled from (0, 0.75). Value (in HSV) is fixed to 1. The sampled HSVs are converted to RGBs.

\textbf{Image.}
An RGB camera is defined in the environment for visual input.
The position of the camera and its facing direction, \emph{i.e.} $(x, y, z, pitch, yaw)$ are defined as actions for agents.
We refer to a viewpoint as the position and facing direction combined.
From a given viewpoint, a generated RGB image has $3 \times 64 \times 64$ dimension.

\textbf{Haptic.} 
For proprioceptive (haptic) sense, the Johns Hopkins Modular Prosthetic Limb (MPL) \citep{johannes2011overview} is used, which is a part of MuJoCo HAPTIX.
The hand model generates 132-dimensional observation, consisting of the its actuator positions, velocities, accelerations, and touch senses.  For more details about the MPL hand, please finds Appendix C. in \cite{amos2018learning}.
The MPL hand model has 13 degrees of freedom to control.
MESE adds 5 degrees of freedom, \emph{i.e.} $(x, y, z, pitch, yaw)$, to control the position and facing direction of the hand's wrist, similar to camera control.

\subsection{Datasets}
Given that each scene has a single object at the origin, images and haptics are randomly generated.
For an image, a camera viewpoint is sampled on a spherical surface with a fixed radius while the camera faces to the object.
We refer to image query as camera viewpoint.

For an haptic data in each scene, we first sample a wrist pose of the hand similar to generating camera viewpoints.
Given the sampled wrist, a fixed deterministic policy is performed.%
The policy starts from a stretched hand pose to gradually go to grabbing posture without any stochasticity.
Note that a haptic datapoint is a function of the wrist pose and the object, given the aforementioned fixed policy; thus, the wrist's position and facing direction is set to the hand's query.
Each dimension of haptic data is re-scaled to $[-1, 1]$.

For the environment $\cM=\{\tt{RGB\textrm{-}image}, \tt{haptics}\}$, also denoted as $|\cM|=2$, 1M scenes are collected for training data.
For each scene, a Shepard-Metzler object with 5 parts is randomly sampled as described in Section \ref{sec:appendix-mese}.
The number of unique shapes is 726 for the 5-parts object dataset.
In each scene, 15 queries and corresponding sensory outputs are randomly sampled for each sensory modularity.
For validation and test data, 20k and 100k scenes are sampled, respectively.

For the environment whose $|\cM|$ is larger than 2, we slice the dimensions of image and haptic data.
For example, in order to build an environment $|\cM|=5$, $3\times64\times64$ image is split to four quadrants of it so that each $3\times32\times32$ image is one of \{$\tt{upper\textrm{-}left\textrm{-}RGB}$, $\tt{upper\textrm{-}right\textrm{-}RGB}$, $\tt{lower\textrm{-}left\textrm{-}RGB}$, $\tt{lower\textrm{-}right\textrm{-}RGB}$\}. In addition to these four visual modalities, haptic input is provided.
Note that while we split each image into four, the corresponding experiment defers from image in-paining or de-noising tasks.
In those image tasks, statistical regularities of image are heavily taken into account, \emph{i.e.} statistics of local receptive fields are almost identical regardless of position.
Many recent solutions on the problems resort to convolutional architectures, as a practical solution for sharing parameters of models across arbitrary locations.
As long as the inductive bias hasn't made use of in any model, it is valid that they are distinct random variables, each of which has different statistical characteristic; thus, they can be treated as multiple modalities.

For $|\cM|=8$, image is cropped to $3\times56\times56$ and re-sized to $3\times48\times48$ due to the memory overhead. The image is split to left-right for each RGB channel; thus, we have \{$\tt{left\textrm{-}R}$, $\tt{left\textrm{-}G}$, $\tt{left\textrm{-}B}$, $\tt{right\textrm{-}R}$, $\tt{right\textrm{-}G}$, $\tt{right\textrm{-}B}$\} as different senses.
Haptic dimension is also divided into to two, \emph{i.e.} $\tt{haptics}_1$ and $\tt{haptics}_2$. $\tt{haptics}_1$ corresponds to thumb, index, and middle fingers. $\tt{haptics}_2$ corresponds to ring and little fingers, as well as palm.

For $|\cM|=14$, image is converted to $3\times48\times48$ as in $|\cM|=8$, but is is sliced as to
$\{\tt{upper, lower}\} \times \{\tt{left, right}\} \times \{\tt{R, G, B}\}$.
With haptic data divded to $\tt{haptics}_1$ and $\tt{haptics}_2$, we get an environment with $|\cM|=14$.

\subsection{Training}
\label{sec:appendix-training}
For training, Adam optimizer is used \citep{KingmaB14}. $\beta$-annealing\footnote{In here, $\beta$-annealing refers to anneal a weight at KL term of ELBO as done in \cite{higgins2017beta}} is employed; $\beta$ is set to 0.1$\sim$1 for the first epoch and maintained as 1 for the rest of training.
Learning rate is set to 0.0001.
In order for stable training, gradient is clipped to $[-0.25, 0.25]$.
Training is ran for 10 epochs.
Mini-batch size is set to 14 scenes for the $|\cM|=2$ environment and 24 scenes for the $|\cM|=$5, 8, and 14.

\section{Network Architectures}
\label{sec:appendix-network-architectures}

\begin{figure}[!hbt]
\centering
\begin{subfigure}[b]{.48\textwidth}
    \centering
    \includegraphics[width=6.cm]{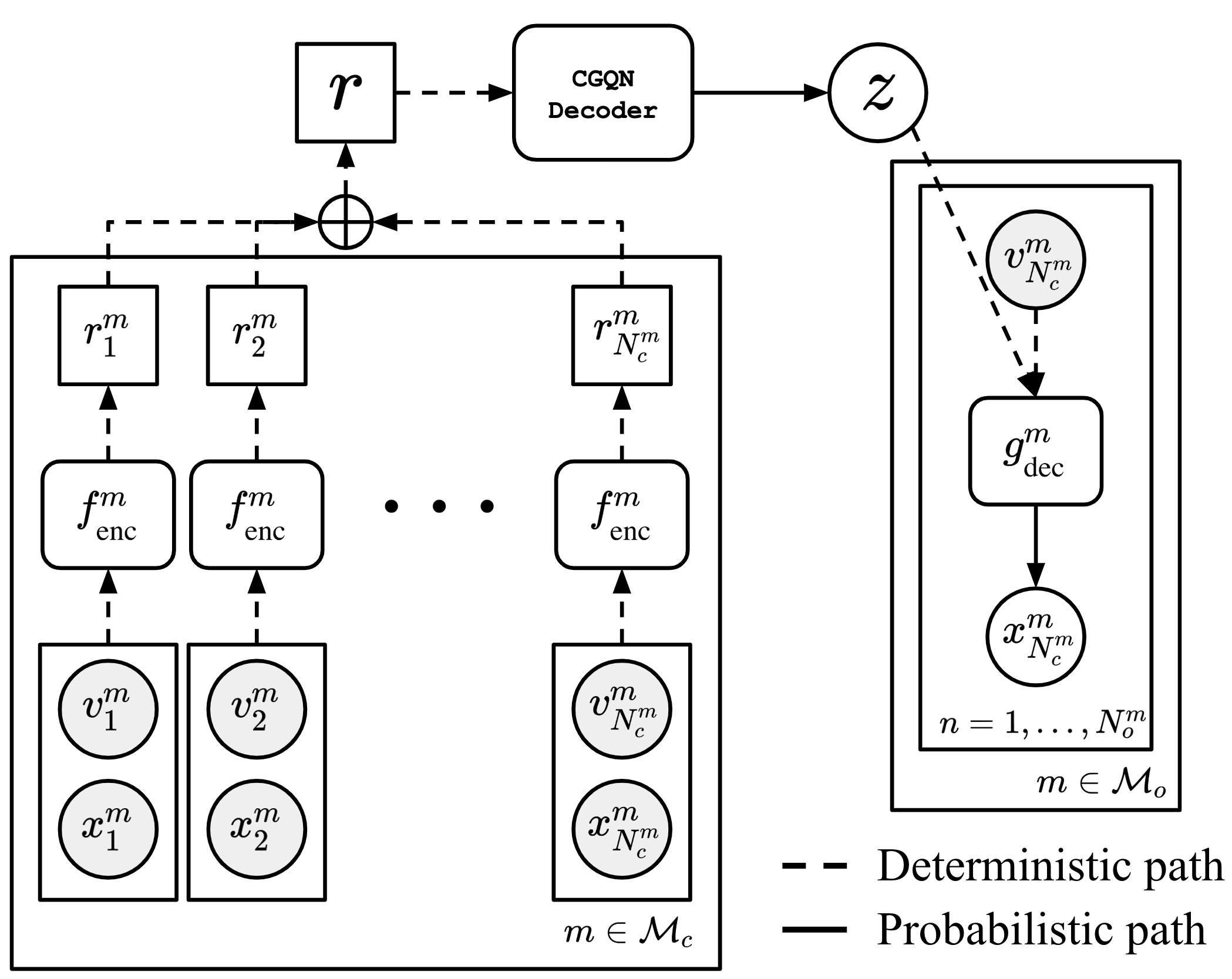}
    \vspace*{0.2cm}
    \caption{
    Baseline model.
    }
    \label{fig:computation-graph-gen-baseline}
\end{subfigure}
\hspace{0.25cm}
\begin{subfigure}[b]{.48\textwidth}
    \centering
    \includegraphics[width=6.cm]{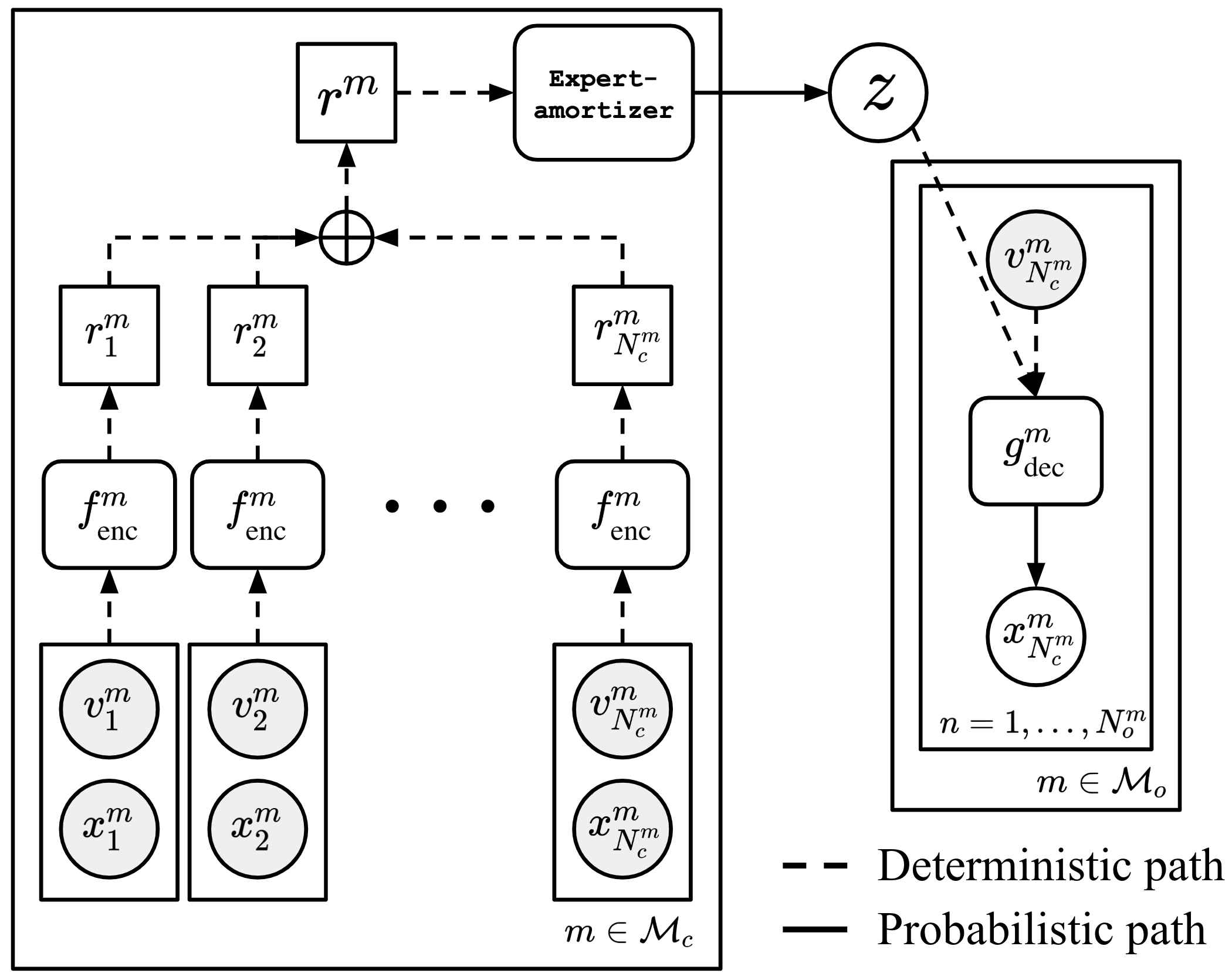}
    \caption{
    APoE.
    }
    \label{fig:computation-graph-gen-APoE}
\end{subfigure}
\caption{
Computation graphs of generation processes for the proposed models.
(a) Baseline model: Each instance of $m$-th modality query-sense pairs feeds to $f_{\text{enc}}^{m}$, \emph{i.e.} representation network.
All instances of representation $\br_n^m$s will summed up to get representation $\br$.
Metamodal scene representation $\bz$ is inferred using the C-GQN decoder (or encoder in inference).
Conditioning on the $\bz$ and a query $\bv_n^m$, an instance of sensory data will be generated using $g_{\text{dec}}^{m}$, \emph{i.e.} the \emph{renderer}.
(b) APoE: Each instance of $m$-th modality query-sense pairs feeds to $f_{\text{enc}}^{m}$, and they are summed up to get modality-specific representation $\br^m$.
Metamodal scene representation $\bz$ is inferred via product-of-experts using the expert-amortizer network.
Rendering follows the same process as in the baseline model.
}
\label{fig:computation-graph-gen}
\end{figure}

\begin{figure}[!hbt]
\centering
\includegraphics[width=6.9cm]{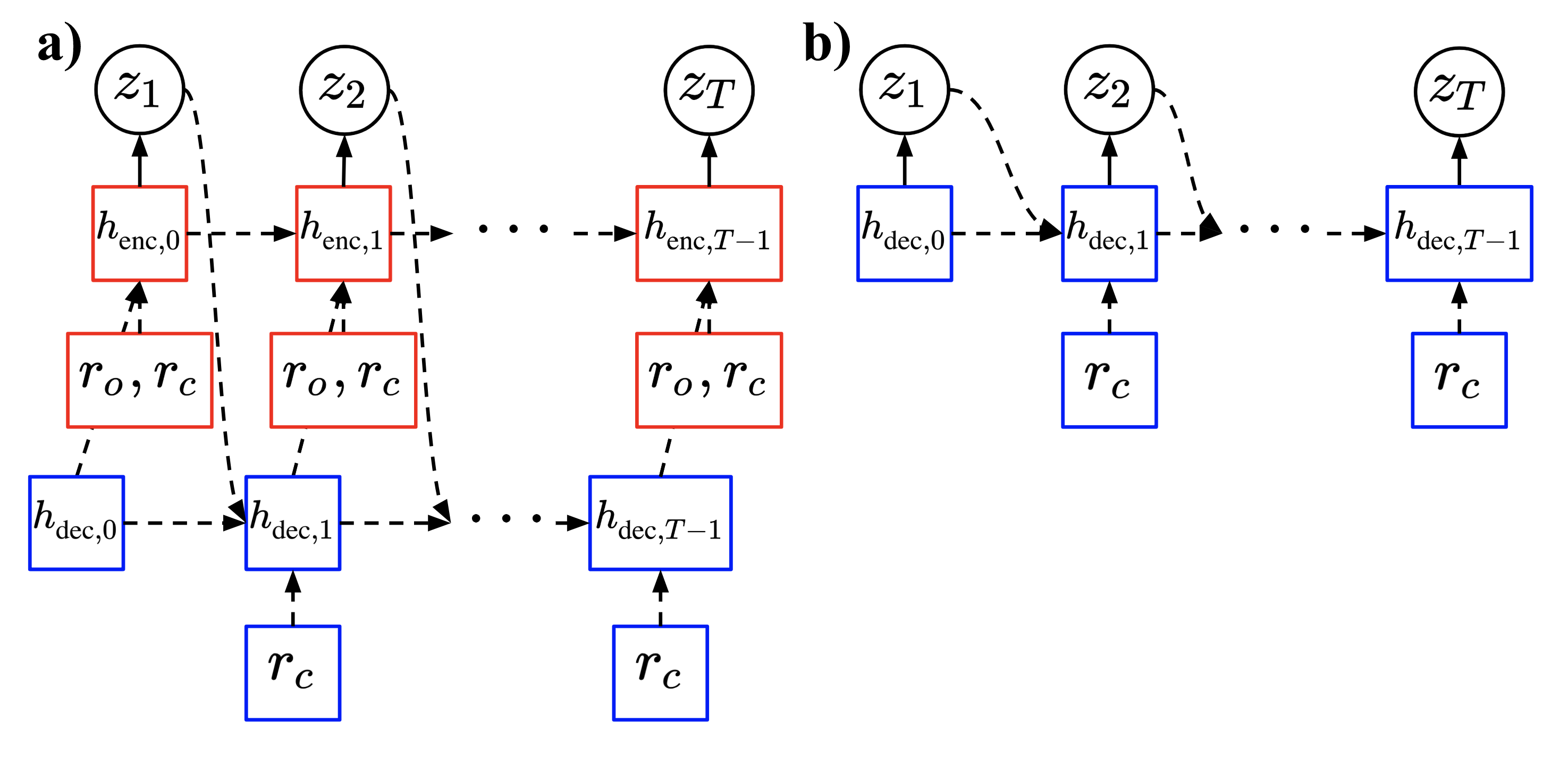}
\includegraphics[width=6.9cm]{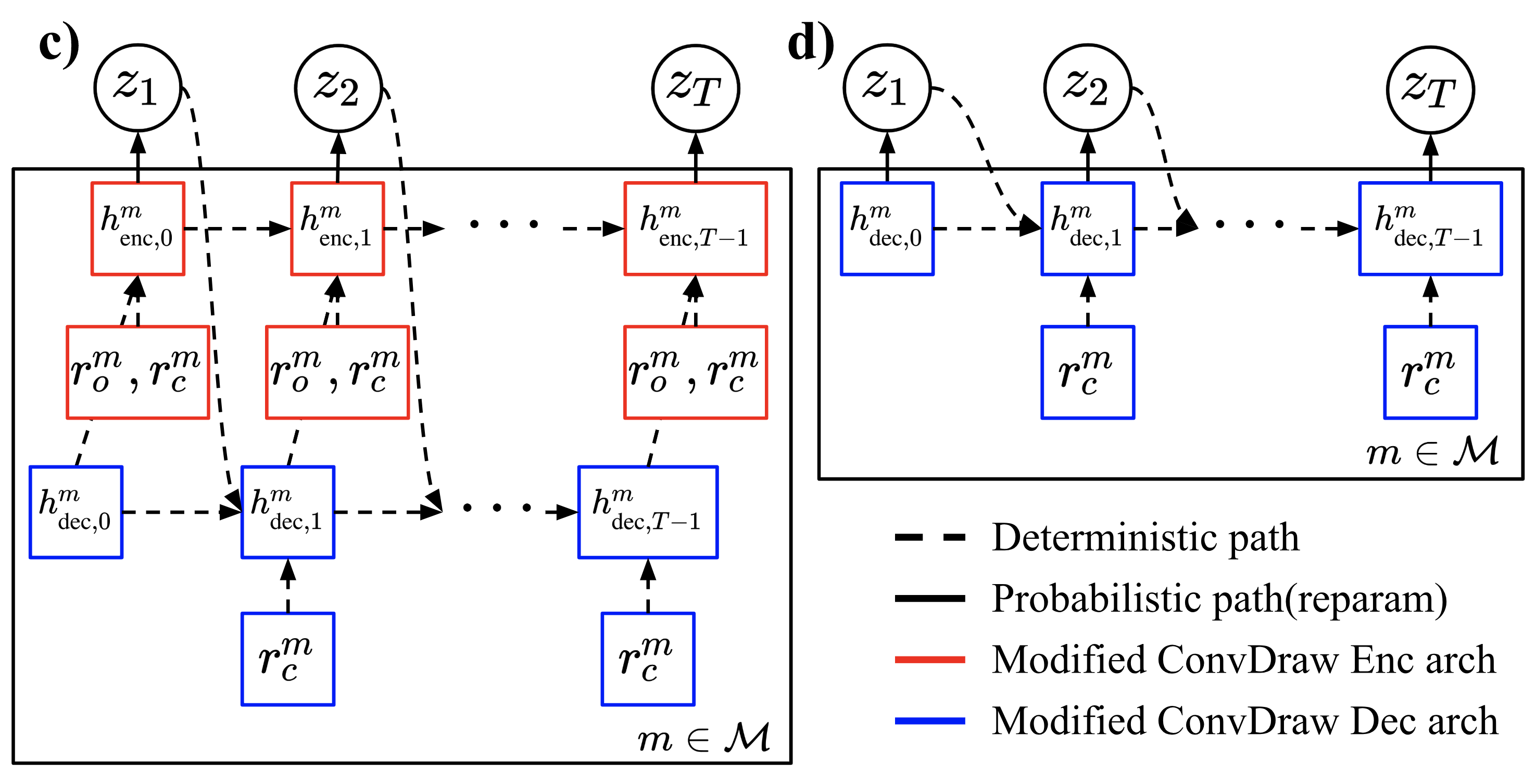}
\caption{
Implementation details for the modified ConvDraw network architecture; a) baseline encoder and b) decoder. c) the proposed model's encoder and d) decoder.
Sampled latent $z=[z_1, z_2, \dots, z_T]$ will be passed to renderers.
Note that in the PoE and APoE, the distribution of $z_t$ is estimated as the product of $m$ experts for each $t$-th step.
}
\label{fig:implementation-baseline-APoE}
\end{figure}

\textbf{Overall.}~We adopt C-GQN network architecture from \cite{KumarERGVLS18} for the proposed model, as well as the baseline.
This architecture can be thought of as a modified version of ConvDraw encoder-decoder, in which the posteriors don't have feedback routes of the predicted inputs and the residuals between the target and the predictions, unlike the original one \citep{GregorBRDW16/nips}.
As a result, for every step of ConvLSTM iterations the same input is repeatedly provided instead (see Fig. \ref{fig:implementation-baseline-APoE} (a)).

For a baseline, we use the C-GQN network, and the baseline's generation process is depicted in Fig. \ref{fig:computation-graph-gen} (a).
Each instance of $m$-th modality query-sense pairs feeds to $f_{\text{enc}}^{m}$, \emph{i.e.} $m$-th representation network.
All instances of representation $\br_n^m$s will summed up to get representation $\br$.
Metamodal scene representation $\bz$ is inferred using the C-GQN decoder (or encoder in inference).
Conditioning on the $\bz$ and a query $\bv_n^m$, a sensory datapoint will be generated using $g_{\text{dec}}^{m}$, \emph{i.e.} the \emph{renderer} for the $m$-th modality.

For APoE, multiple experts are modeled as a single network, called expert-amortizer, in which a binary mask to identify modularity is used while inferring $z$, e.g. $Q_{\phi_m}(\bz|\br^m) = Q_{\phi}(\bz|\br^m, m) \textrm{ for } \forall m \in \cM$ in Eq. \eqref{eq:posterior-APoE} where $m=$ a binary mask.
The expert-amortizer is build upon further modifications from the modified ConvDraw, as shown in Fig. \ref{fig:implementation-baseline-APoE} (b).
Especially for efficient computation, the expert-amortizers are implemented such that they perform convolution over $\br^m \textrm{ for } \forall m \in \cM_S$.

See Fig.~\ref{fig:computation-graph-gen}~(b) for APoE's generation.
Identical to the baseline, each instance of $m$-th modal query-sense pairs feeds to $f_{\text{enc}}^{m}$, and they are summed up to get modality-specific representation~$\br^m$.~However, metamodal scene representation $\bz$ is inferred via product-of-experts using the expert-amortizer network.

For PoE, each expert is modeled a single ConvDraw encoder-decoder with corresponding modularity encoder, and the rest of its implementations are identical to APoE.

\textbf{Representation Network.}~To estimate modality-specitic representation for each instance of a query-sense pair, 
tower representation networks proposed in \cite{Eslami1204/science18} is used.
For camera position-image pair, convolution layer is used in the tower representation network.
Similar to image, MLP is applied for a haptic observation and its corresponding query.
The same representation network architectures are used to baseline, PoE, and APoE.

\textbf{Renderer.}~Renderer network is a part of a decoder for predicting each sensory modality.
Those renderers get a query $\bv_m^n$ and modality-agnostic latent representation $z$ as its inputs, output the sensory data conditioning on them.
For rendering image, ConvLSTM with convolutional layer is used as done in C-GQN.
Similar to image rendering, ConvLSTMs is used for proprioceptive, but MLP is employed instead of convolution layer.

\section{Classification}
\label{sec:appendix-classification}
For classification, we adopt a method from \cite{lake2015human/science}.
Let we have $K$-number of context sets, $C^{(k)}$ for $\forall$ $k$, each of which is a set that contains multisensory data-query pairs.
Given we have an observation set $O=\{X, V\}$ obtained from one of the $K$-scenes (more precisely objects), we can predict from which scene the new observation set comes. The predicted label $\hat{k}$ is obtained by following;
\eq{
\label{eq:classification}
\hat{k} = \argmax_{k} \log P_\ta(X|V,C^{(k)})
,
}
where each $\log P_\ta(X|V,C^{(k)})$ is estimated by using the log-likelihood estimators from \cite{BurdaGS15/iclr}. This method doesn't require additional training process. To approximate the log-likelihood each $k$, 50 latent samples are used.

For held-out dataset, 1000 additional Shepard-Metzler objects with 4 or 6 parts are generated: any of these objects hasn't presented in training dataset.
$K$ is set to 10.
For all models, three different inference scenarios are considered; classification is performed by using (i) only image-query pairs from each scene ($I$), (ii) haptic-only contexts ($H$), and (iii) use both sensory contexts ($H+I$).

\begin{figure}[!h]
\centering
\includegraphics[width=5.88cm]{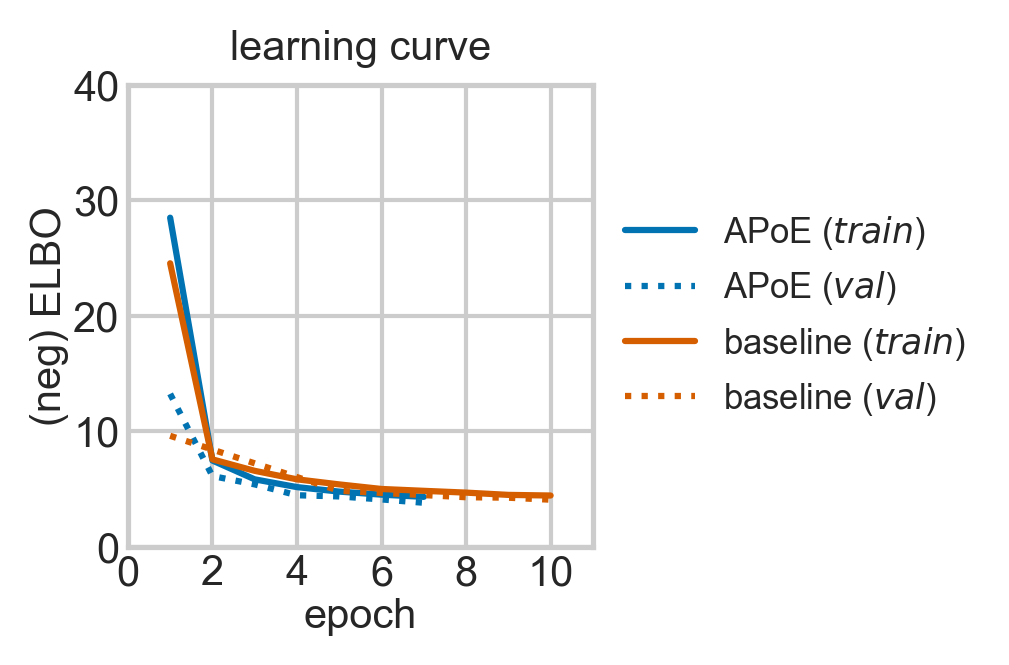}
\includegraphics[width=6cm]{figs/Fig9_clsinf_gray_img_hpt_m2.png}
\caption{
Classification result for $|\cM|=2$ environment.
(Left) learning curves for baseline and APoE models used in classification.
(Right) classification results of the models from the left.
Classification is performed in three different conditioning scenarios; each model is conditioned on image-only context ($I$), haptic-only context ($H$) and all information ($H+I$).
}
\label{fig:result-elbo-classification-m2}
\end{figure}

The results are shown in \ref{fig:result-elbo-classification-m2}.
In order to claim that both models are well trained and converged to training dataset, the learning curves for baseline and APoE models used in classification are also attached.

\section{Cross-modal Generation}
\label{sec:appendix-cross-modal-generation}

\subsection{Reducing Uncertainty with Aggregation of Evidences}
In this task, we examine the uncertainty of modality-agnostic representations with respect to the number of contexts.
Similar to Fig.~\ref{fig:result-cross-modal-generation}, we provide a single image context but we condition a trained model on different numbers of haptic contexts.
More precisely, the image context is given such that the model cannot recover the entire scene from the image.

\begin{figure}[!h]
\centering
\begin{subfigure}{\textwidth}
    \centering
    \begin{subfigure}{0.48\textwidth}
        \centering
        \includegraphics[width=\textwidth]{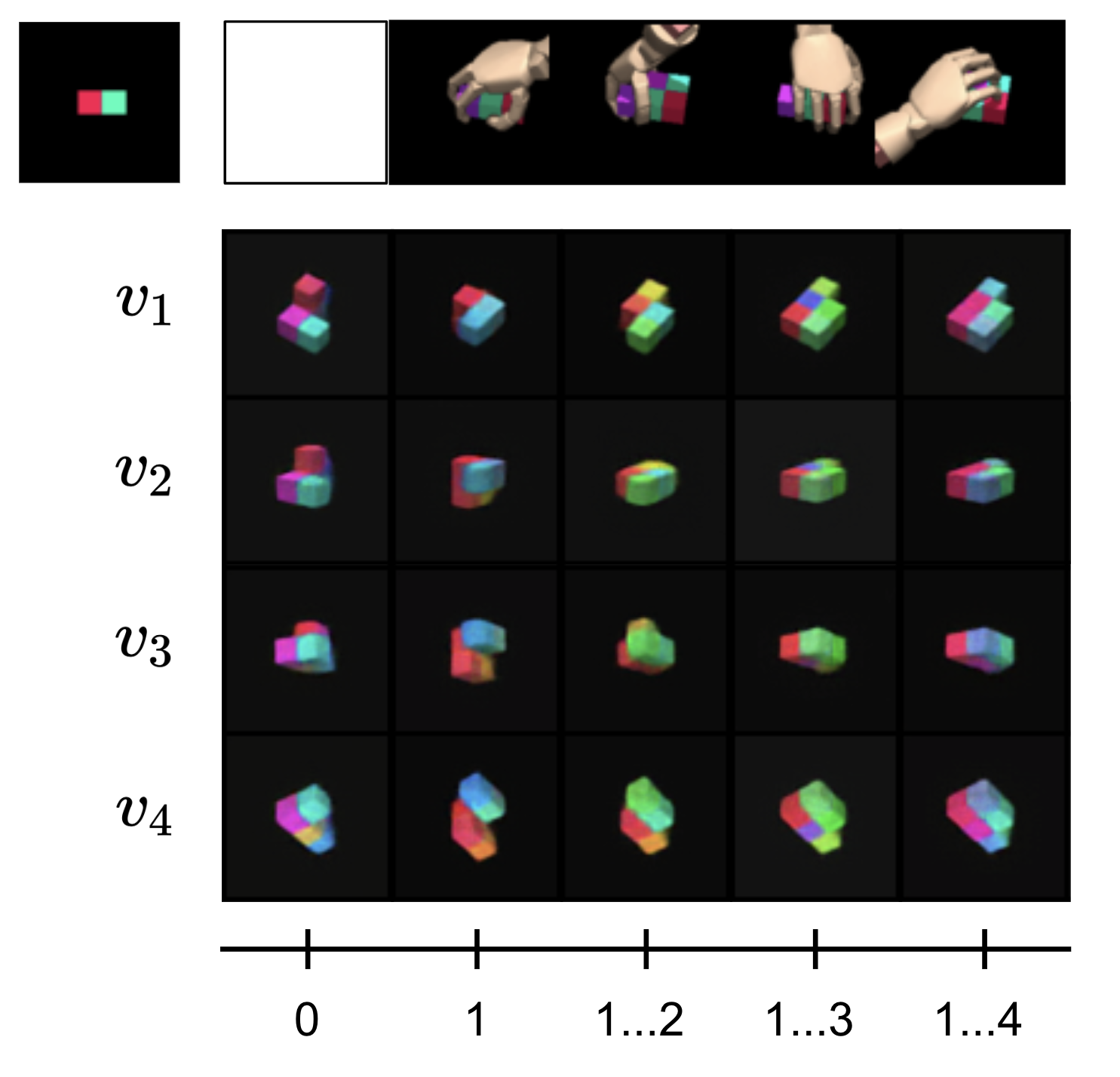}
        \vspace{-0.5cm}
        \caption{}
    \end{subfigure}
    \begin{subfigure}{0.096\textwidth} %
        \centering
        \includegraphics[width=\textwidth]{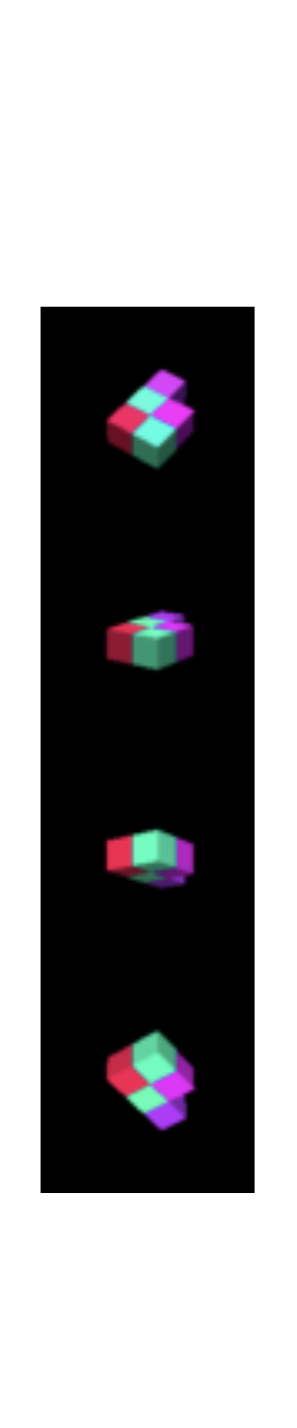}
        \vspace{-0.5cm}
        \caption{}
    \end{subfigure}
    \vspace*{0.2cm}
\end{subfigure}
\begin{subfigure}{\textwidth} %
    \centering
    \includegraphics[width=\textwidth]{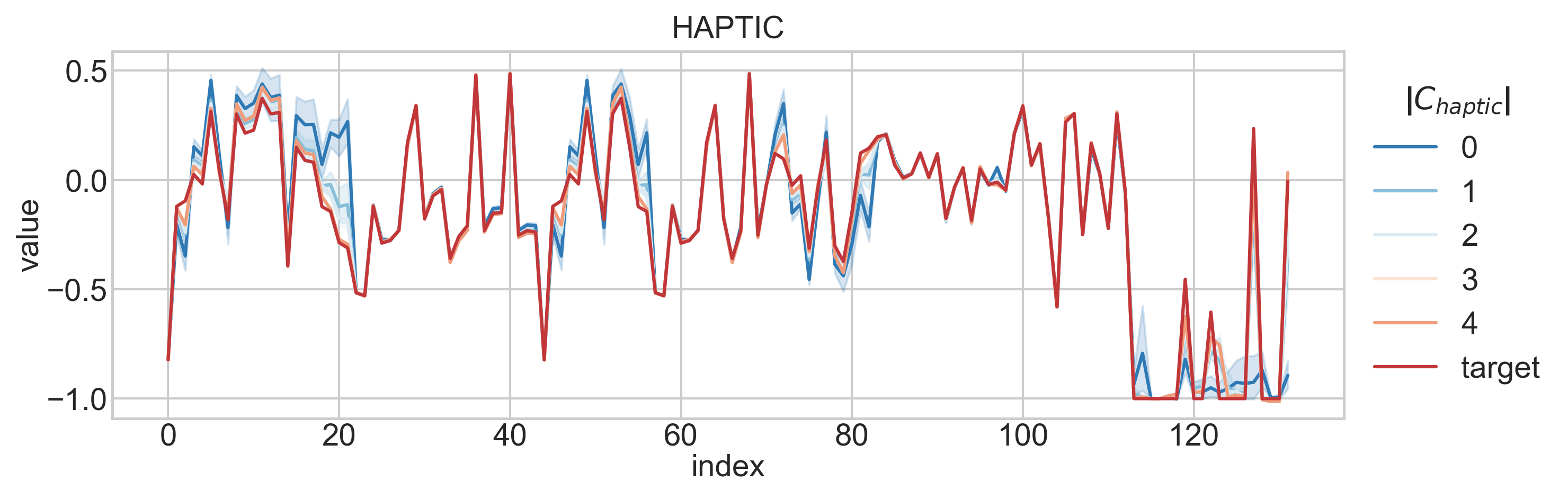}
    \vspace{-0.5cm}
    \caption{}
\end{subfigure}
\caption{
Multi-sensory inference.
(a) Upper row illustrates single visual context and various haptic cues, from empty observation to multiple observations.
Note that the given visual context is insufficient to infer correct object shape.
The model predicts visual observations for different viewpoints, \emph{i.e.} $v_1$, $v_2$, and $v_3$, using the visual and haptic contexts.
$x$-axis label indicates the indices of haptic contexts used when the model predicts the corresponding column.
(b) The ground truth images for the same viewpoints.
(c) The model predicts haptic observations for a haptic query.
The ground truth values are denoted as red. $x$-axis label indicates index of 132-dimensional output of the hand model.
$|C_{\tt{haptics}}|$ means the number of haptic contexts used for prediction.
}
\label{fig:result-reducing-uncertainity-with-aggregration}
\end{figure}

The generated image samples are shown in Fig. \ref{fig:result-reducing-uncertainity-with-aggregration} (a).
As the number of the haptic contexts increases, more accurate visual observation is predicted.
We can also observe that generated images at each column continue to develop in comparison to the previous column, corresponding to where additional haptic information is provided.
Again, we observe that the part of the object for which the context image provides color information has similar colors while other part of the block has random colors.

Fig. \ref{fig:result-reducing-uncertainity-with-aggregration} (b) describes the generated haptic samples from the same query.
In this figure, 95\%-confidence interval from 20 samples is also illustrated.
Similar to visual prediction, the haptic prediction improves according to the number of the haptic contexts.
In addition, it is demonstrated that the uncertainty of the prediction reduces as the contexts aggregate.

\subsection{Any-to-any Cross-modal Generation}
\label{sec:appendix-any-to-any-generation}

\begin{figure}[!ph]
\centering
\begin{subfigure}{\textwidth}
    \centering
    \includegraphics[width=3.5cm]{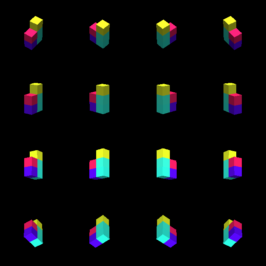}
    \vspace{-0.15cm}
    \caption{
    Ground-truth images
    }
    \vspace*{0.2cm}
\end{subfigure}
\begin{subfigure}{\textwidth}
    \centering
    \begin{subfigure}[t]{.3\textwidth}
        \centering
        \includegraphics[width=3.5cm]{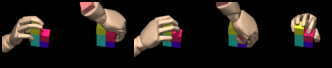}
        \includegraphics[width=3.5cm]{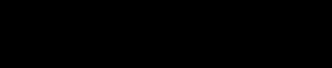}
        \includegraphics[width=3.5cm]{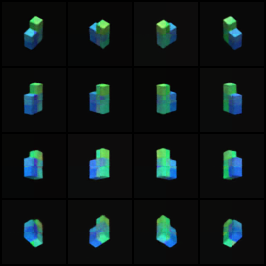}
    \end{subfigure}
    \begin{subfigure}[t]{.3\textwidth}
        \centering
        \includegraphics[width=3.5cm]{figs-gen-c72/m5/ctx-hand-img-cls72-i1-nc5.png}
        \includegraphics[width=3.5cm]{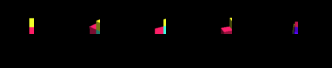}
        \includegraphics[width=3.5cm]{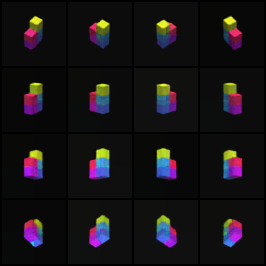}
    \end{subfigure}
    \begin{subfigure}[t]{.3\textwidth}
        \centering
        \includegraphics[width=3.5cm]{figs-gen-c72/m5/ctx-hand-img-cls72-i1-nc5.png}
        \includegraphics[width=3.5cm]{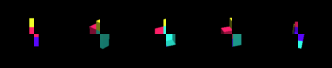}
        \includegraphics[width=3.5cm]{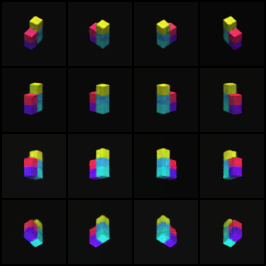}
    \end{subfigure}
    \vspace{-0.15cm}
    \caption{
    $|\cM|=5$ environment, \emph{i.e.} $\{\tt{upper, lower}\} \times \{\tt{left, right}\} + \tt{haptics}$.
    }
    \vspace*{0.2cm}
\end{subfigure}
\begin{subfigure}{\textwidth}
    \centering
    \begin{subfigure}[t]{.3\textwidth}
        \centering
        \includegraphics[width=3.5cm]{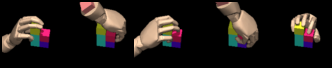}
        \includegraphics[width=3.5cm]{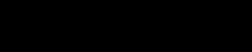}
        \includegraphics[width=3.5cm]{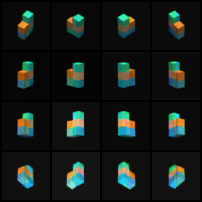}
    \end{subfigure}
    \begin{subfigure}[t]{.3\textwidth}
        \centering
        \includegraphics[width=3.5cm]{figs-gen-c72/m8/ctx-hand-img-cls72-i1-nc5.png}
        \includegraphics[width=3.5cm]{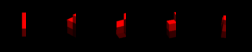}
        \includegraphics[width=3.5cm]{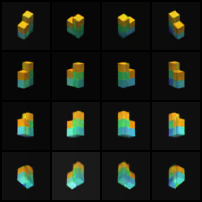}
    \end{subfigure}
    \begin{subfigure}[t]{.3\textwidth}
        \centering
        \includegraphics[width=3.5cm]{figs-gen-c72/m8/ctx-hand-img-cls72-i1-nc5.png}
        \includegraphics[width=3.5cm]{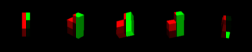}
        \includegraphics[width=3.5cm]{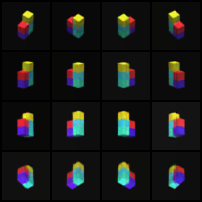}
    \end{subfigure}
    \vspace{-0.15cm}
    \caption{
    $|\cM|=8$ environment, \emph{i.e.} $\{\tt{left, right}\} \times \{\tt{R, G, B}\} + \tt{haptics}_1 + \tt{haptics}_2$.
    }
    \vspace*{0.2cm}
\end{subfigure}
\begin{subfigure}{\textwidth}
    \centering
    \begin{subfigure}[t]{.3\textwidth}
        \centering
        \includegraphics[width=3.5cm]{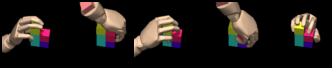}
        \includegraphics[width=3.5cm]{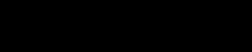}
        \includegraphics[width=3.5cm]{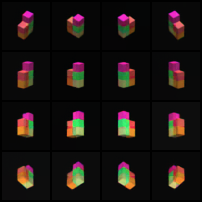}
    \end{subfigure}
    \begin{subfigure}[t]{.3\textwidth}
        \centering
        \includegraphics[width=3.5cm]{figs-gen-c72/m14/ctx-hand-img-cls72-i1-nc5.png}
        \includegraphics[width=3.5cm]{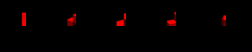}
        \includegraphics[width=3.5cm]{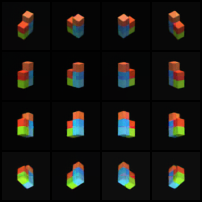}
    \end{subfigure}
    \begin{subfigure}[t]{.3\textwidth}
        \centering
        \includegraphics[width=3.5cm]{figs-gen-c72/m14/ctx-hand-img-cls72-i1-nc5.png}
        \includegraphics[width=3.5cm]{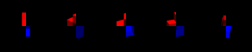}
        \includegraphics[width=3.5cm]{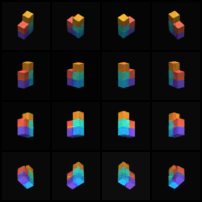}
    \end{subfigure}
    \vspace{-0.15cm}
    \caption{
    $|\cM|=14$ environment, \emph{i.e.} $\{\tt{upper, lower}\} \times \{\tt{left, right}\} \times \{\tt{R, G, B}\} + \tt{haptics}_1 + \tt{haptics}_2$.
    }
\end{subfigure}
\caption{
Any-to-any cross-modal generation examples.
Given context conditions, trained models are asked to generate all modality outputs (for a given set of queries) and these are combined to be displayed.
The first row in each sub-figure displays five haptic context senses.
The second row illustrates combined senses from different visual modalities.
The third illustrates the predictions for given queries.
For example, the second row in (b) depicts the five $\tt{upper\textrm{-}left\textrm{-}RGB}$ senses. The same row in (c) displays additional five $\tt{lower\textrm{-}right\textrm{-}RGB}$ senses combined with the ones in (b).
}
\label{fig:result-cross-modal-generation-extended}
\end{figure}

Additional cross-modal generation experiments are performed for $|\cM|=5, 8, \text{and}, 14$ in order to explore multisensory integration of arbitrary context conditions.
Given any context condition, a trained model is asked to generate all modality outputs (for a given set of queries) and these are combined to be displayed.
For instance, a model trained in $|\cM|=14$ generates outputs in all modalities, \emph{i.e.} $\cM = \{\tt{upper, lower}\} \times \{\tt{left, right}\} \times \{\tt{R, G, B}\} + \tt{haptics}_1 \tt{haptics}_2$.
The visual outputs are combined and displayed as shown in Fig \ref{fig:result-cross-modal-generation-extended} (d).
The haptic outputs are omitted to conserve space.

Three different context conditions are applied for each environment.%
For $|\cM|=5$, a model is conditioned on (i) $\tt{haptics}$-only, (ii) $\tt{haptics} + \tt{upper\textrm{-}left\textrm{-}RGB}$, and (iii) $\tt{haptics} + \tt{upper\textrm{-}left\textrm{-}RGB} + \tt{lower\textrm{-}right\textrm{-}RGB}$ contexts.
For $|\cM|=8$, a model is conditioned on (i) $\tt{haptics}_1 + \tt{haptics}_2$, (ii) $\tt{haptics}_1 + \tt{haptics}_2 + \tt{left\textrm{-}R}$, and (iii) $\tt{haptics}_1 + \tt{haptics}_2 + \tt{left\textrm{-}R} + \tt{right\textrm{-}G}$.
For $|\cM|=14$, a model is conditioned on (i) $\tt{haptics}_1 + \tt{haptics}_2$, (ii) $\tt{haptics}_1 + \tt{haptics}_2 + \tt{upper\textrm{-}left\textrm{-}R}$, and (iii) $\tt{haptics}_1 + \tt{haptics}_2 + \tt{upper\textrm{-}left\textrm{-}R} + \tt{lower\textrm{-}right\textrm{-}B}$.
Each context modality is provided with 5 query-sense pairs.

The results are shown in Fig. \ref{fig:result-cross-modal-generation-extended} (b)-(d), and the ground truth images are given in Fig. \ref{fig:result-cross-modal-generation-extended} (a).
The provided context senses are illustrated in the first and second rows in each experiment.
In general, haptic-related contexts are sufficient for the learned models to infer the shapes.
With additional visual cues, the models start to correctly predict colors.
For example, in middle column of Fig. \ref{fig:result-cross-modal-generation-extended} (c) and (d), red-mixed colors are successfully inferred with $R$-channel context; however, it still fails to predict all color patterns as other color information is deficient.
As more color information is given, our models results in successful predictions of all color patterns as shown in the right column of Fig. \ref{fig:result-cross-modal-generation-extended} (c) and (d).

\section{Missing-modality Problem}
\label{sec:appendix-incomplete-multisensory-data}

\begin{figure}[!ph]
\centering
\begin{subfigure}{\textwidth}
    \centering
    \includegraphics[width=3.10cm]{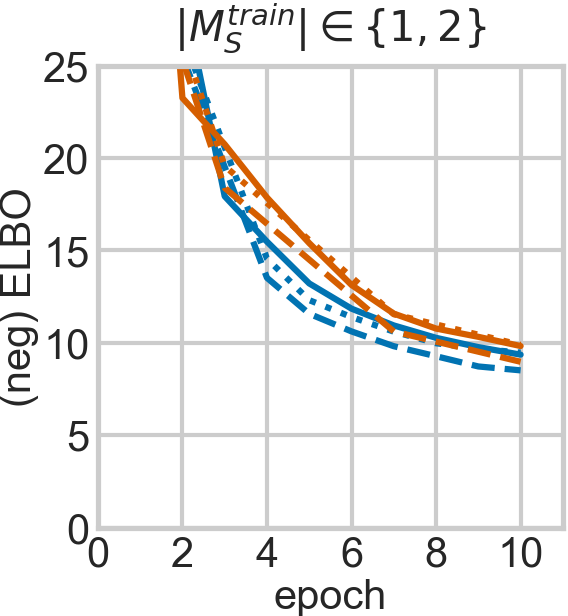}
    \includegraphics[width=2.63cm]{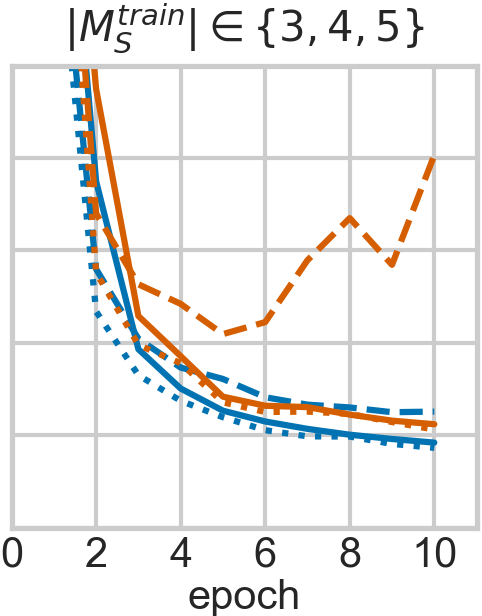} %
    \includegraphics[width=5.06cm]{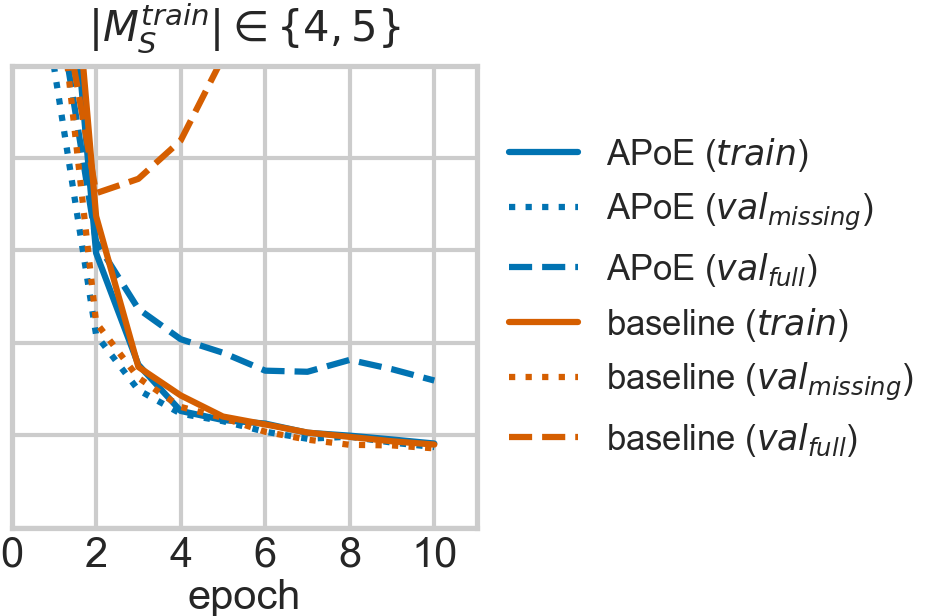} %
    \caption{
    $|\cM|=5$ environment, \emph{i.e.} $\{\tt{upper, lower}\} \times \{\tt{left, right}\} + \tt{haptics}$.
    }
    \vspace*{1cm}
\end{subfigure}
\begin{subfigure}{\textwidth}
    \centering
    \includegraphics[width=3.10cm]{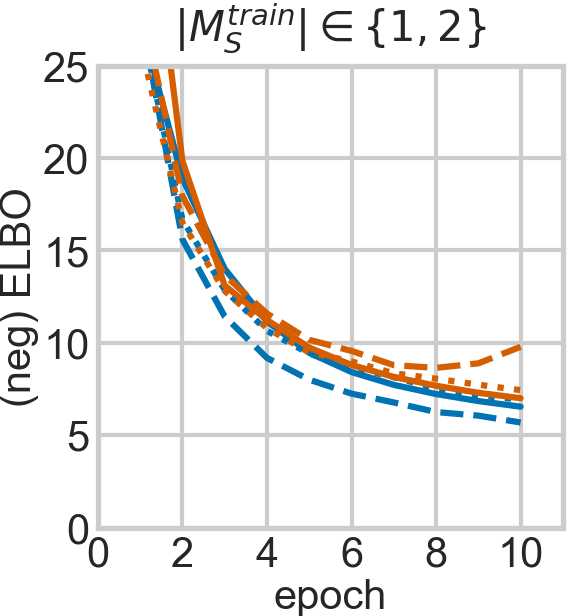}
    \includegraphics[width=2.63cm]{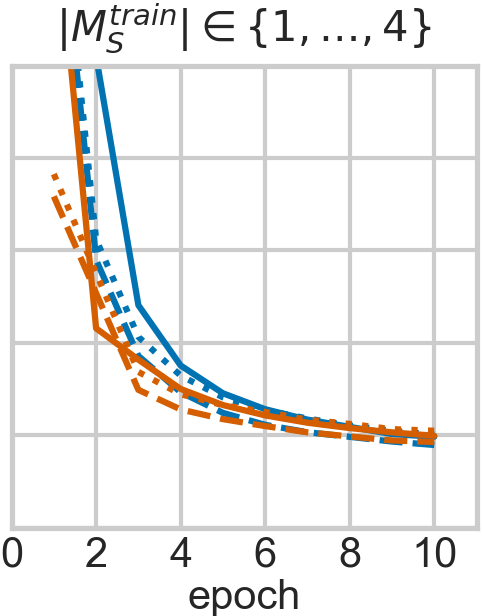} %
    \includegraphics[width=2.63cm]{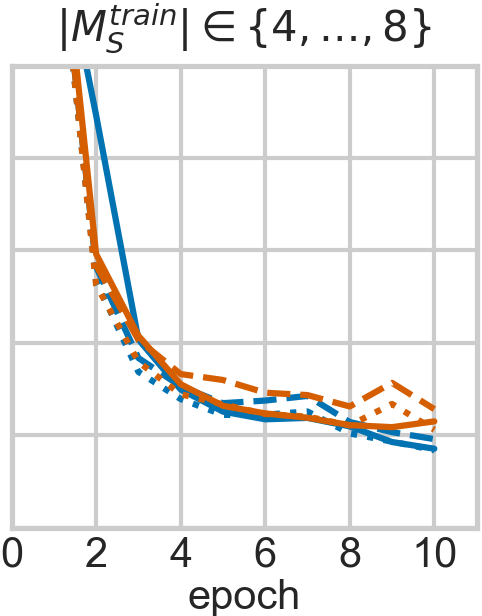} %
    \includegraphics[width=5.06cm]{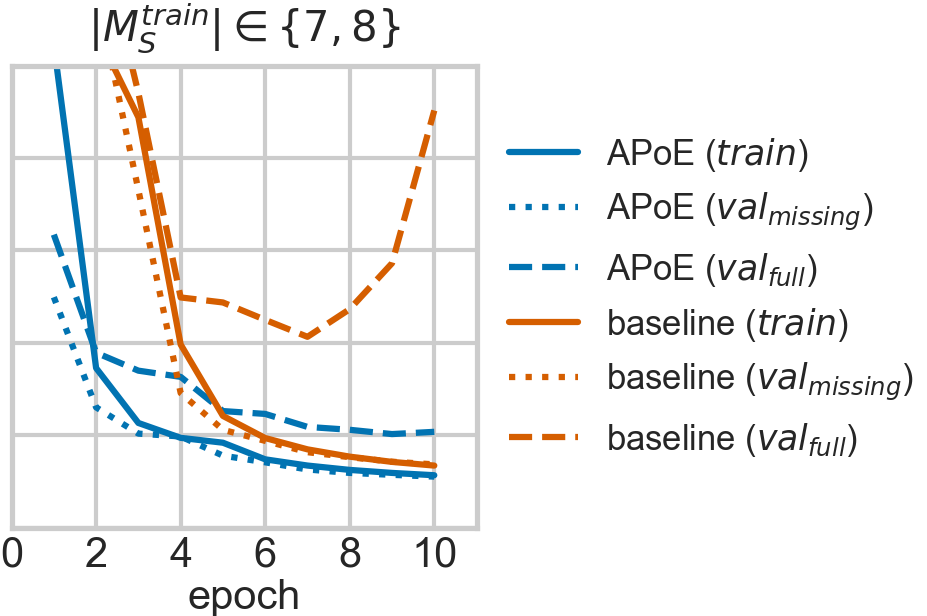} %
    \caption{
    $|\cM|=8$ environment, \emph{i.e.} $\{\tt{left, right}\} \times \{\tt{R, G, B}\} + \tt{haptics}_1 + \tt{haptics}_2$.
    }
    \vspace*{1cm}
\end{subfigure}
\begin{subfigure}{\textwidth}
    \centering
    \includegraphics[width=3.10cm]{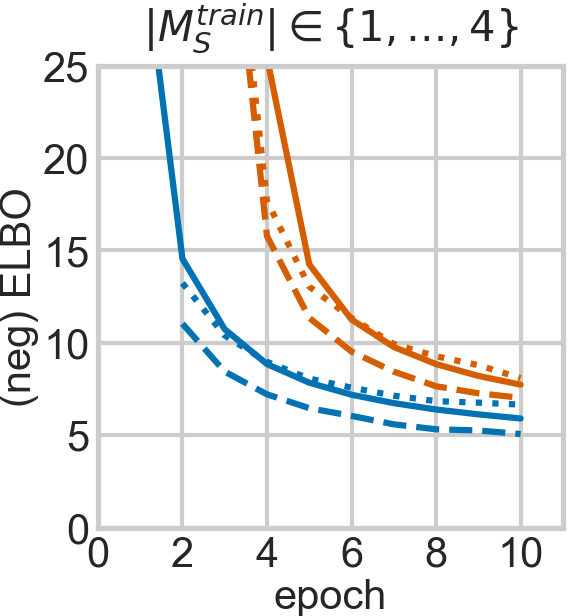}
    \includegraphics[width=2.63cm]{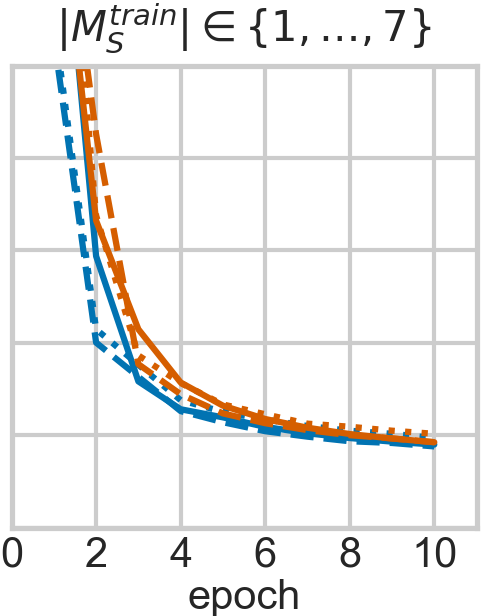} %
    \includegraphics[width=2.63cm]{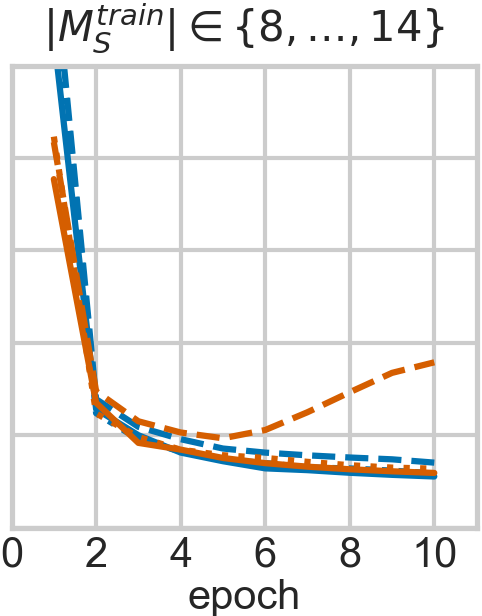} %
    \includegraphics[width=5.06cm]{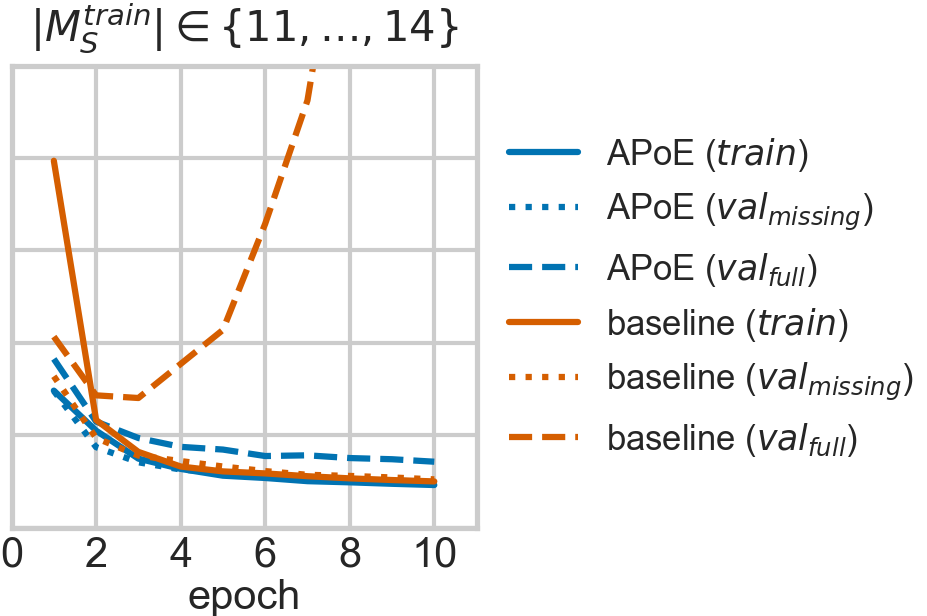} %
    \caption{
    $|\cM|=14$ environment, \emph{i.e.} $\{\tt{upper, lower}\} \times \{\tt{left, right}\} \times \{\tt{R, G, B}\} + \tt{haptics}_1 + \tt{haptics}_2$.
    }
    \vspace*{1cm}
\end{subfigure}
\caption{
Results of missing-modality experiments for various multimodal scenarios.
During training, limited combinations of modalities are presented.
At test time, all combinations of the entire modalities are randomly selected.
}
\label{fig:result-extrapolation-interpolation-extended}
\end{figure}

In addition to the experiments described in Fig. \ref{fig:result-extrapolation-interpolation}, more results are added in \ref{fig:result-extrapolation-interpolation-extended}.
Note that $\tt{train}$ loss is evaluated as moving average of mini-batches, while ${\tt{val}}_{\tt{missing}}$ is estimated on whole batch at the end of each epoch.
This explains validation loss sometimes lower than training's in the figures.

In general, all models tend to under-fit when they have never seen entire modalities during training.
On the other hand, the models exposed to many modalities are prone to give us tighter negative ELBO.

We can observe notable difference between the baseline and APoE on the settings secluded from individual modality during training.
Combined with the classification in Fig. \ref{fig:result-elbo-classification-m2}, we can interpret the results as that PoE helps training individual expert.
The inference of PoE has been understood as an agreement of all experts \citep{Hinton02/neco}; therefore, this lead each expert is capable of performing inference independently as well as expressing its own uncertainty. 
On the other hand, the simple sum operation of the C-GQN (baseline) probably end up with relying on dominating signals and ignore rests, which drove to overfit to training distributions.

\section{Computational Time}

\begin{table}[!h]
    \centering
    \begin{tabular}{l c c c c | c c c c}
        \toprule
        \multirow{2}{*}{Model} & \multicolumn{4}{c|}{\# of parameters} & \multicolumn{4}{c}{timer per iter (ms)} \\ \cmidrule{2-9} %
                               & $|\cM|=$2       & 5       & 8       & 14     & $|\cM|=$2        & 5        & 8       & 14      \\
        \midrule
        baseline               & 53M     & 28M     & 48M     & 51M    & 346      & 397      & 481      & 866     \\
        APoE                   & 53M     & 29M     & 48M     & 51M    & 587      & 679      & 992      & 1189    \\ 
        PoE                    & 58M     & 53M     & 92M     & 131M   & 486      & 790      & 1459     & 2059    \\
        \bottomrule
    \end{tabular}
    \vspace*{0.25cm}
    \caption{
    The number of parameters and computation time for each model for all experiments. Mini batch size is set to 1.
    }
    \label{tab:computation-costs}
\end{table}

Table~\ref{tab:computation-costs} shows the number of parameters and computational time cost for all experiments.
Each experiment is ran with single NVIDIA Tesla P100 GPU and four cores of an Intel Xeon E5-2650 2.20GHz CPU.
PyTorch~\citep{paszke2017pytorch}, CUDA-9.0~\citep{nickolls2008cuda}, and cuDNN7~\citep{chetlur2014cudnn} are used for the implementations.
All models share the same representation and renderer network architectures, and the same number of steps and hidden sizes are applied to the encoder and decoder architectures.
For fair comparison, the mini-batch size is set to 1 for measuring the costs.

In PoE, each of the expert contains a large network like ConvDraw, resulting in $O(|M|)$ space cost for inference networks.
In APoE, the inference networks are integrated into single expert-amortizer, serving for all modalities. Thus, the space cost of inference networks reduces to $O(1)$.
As a result, the APoE model's parameter size in the experiments is almost the same as the baseline's, while it can provide probabilistic information integration that the PoE has.

\newpage
\null
\vfill
\vfill

\end{document}